\documentclass[11pt, letterpaper]{article}
\usepackage[utf8]{inputenc}
\usepackage[margin=1.5cm]{geometry}
\usepackage{titlesec}
\usepackage{tabu}
\usepackage{xcolor}
\usepackage{booktabs}
\usepackage[hyphens]{url}
\usepackage{array}
\usepackage{makecell}
\usepackage{enumitem}
\usepackage{amssymb}
\usepackage{eurosym}
\newlist{selectlist}{itemize}{2}
\setlist[selectlist]{label=$\square$,leftmargin=*,noitemsep,topsep=0pt}
\usepackage{lmodern}
\usepackage{hyperref}
\hypersetup{
    colorlinks=true,
    linkcolor=blue,
    filecolor=magenta,
    urlcolor=cyan,
}
\titleformat{\section}[block]{\hspace{1em}\bfseries}{\thesection.}{0.5em}{} 
\titleformat{\subsection}[block]{\hspace{1em}}{\thesubsection}{0.5em}{}
\titleformat{\subsubsection}[block]{\hspace{1em}}{\thesubsubsection}{0.5em}{}
\RequirePackage{graphicx}
\RequirePackage{float}
\RequirePackage{subfigure}
\RequirePackage{fancybox}
\RequirePackage{color}
\RequirePackage{balance}
\usepackage{xcolor,colortbl}
\usepackage{listings}
\definecolor{codegreen}{rgb}{0,0.6,0}
\definecolor{codegray}{rgb}{0.5,0.5,0.5}
\definecolor{codepurple}{rgb}{0.58,0,0.82}
\definecolor{backcolour}{rgb}{0.95,0.95,0.92}
\lstdefinestyle{mystyle}{
    backgroundcolor=\color{backcolour},   
    commentstyle=\color{codegreen},
    keywordstyle=\color{magenta},
    stringstyle=\color{codepurple},
    basicstyle=\ttfamily\footnotesize,
    breakatwhitespace=false,         
    breaklines=true,                 
    captionpos=b,                    
    keepspaces=false,                 
    showspaces=false,                
    showstringspaces=false,
    showtabs=false,                  
    tabsize=2
}
\begin{document}

\setlength{\parindent}{0pt}
\setlength{\parskip}{10pt}

\textbf{Article title}\\
$ROMR$: A ROS-based Open-source Mobile Robot

\textbf{Authors}\\
Nwankwo Linus*, Fritze Clemens, Konrad Bartsch, Elmar Rueckert

\textbf{Affiliations}\\
Chair of Cyber-Physical Systems, Montanuniversität, 8700 Leoben, Austria

\textbf{Corresponding author’s email address}\\
linus.nwankwo@unileoben.ac.at

\textbf{Abstract}\\
Currently, commercially available intelligent transport robots that are capable of carrying up to 90kg of load can cost \$5,000 or even more. This makes real-world experimentation prohibitively expensive and limits the applicability of such systems to everyday home or industrial tasks. Aside from their high cost, the majority of commercially available platforms are either closed-source, platform-specific or use difficult-to-customize hardware and firmware. In this work, we present a low-cost, open-source and modular alternative, referred to herein as "\underline{R}OS-based \underline{O}pen-source \underline{M}obile \underline{R}obot ($ROMR$)". $ROMR$ utilizes off-the-shelf (OTS) components, additive manufacturing technologies, aluminium profiles, and a consumer hoverboard with high-torque brushless direct current (BLDC) motors. $ROMR$ is fully compatible with the robot operating system (ROS), has a maximum payload of 90kg, and costs less than \$1500. Furthermore, $ROMR$ offers a simple yet robust framework for contextualizing simultaneous localization and mapping (SLAM) algorithms, an essential prerequisite for autonomous robot navigation. The robustness and performance of the $ROMR$ were validated through real-world and simulation experiments. All the design, construction and software files are freely available online under the GNU GPL v3 license at \url{https://doi.org/10.17605/OSF.IO/K83X7}. A descriptive video of $ROMR$ can be found at \url{https://osf.io/ku8ag}.


\arrayrulecolor{black}
\begin{table}[H]
\caption{Specification table}
\begin{center}
\begin{tabular}{p{0.25\linewidth} p{0.50\linewidth}}
  \toprule
    \textbf{Hardware name} & A ROS-based open-source mobile robot ($ROMR$)\\
    \textbf{Subject area} & 
  \begin{itemize}[noitemsep, topsep=0pt]
  \item Robotics
  \item Sensor fusion
  \item Simultaneous localisation and mapping (SLAM)
  \item Navigation
  \item Teleoperation
  \item Research and development in robotics
  \item General
  \end{itemize}  \\
 \textbf{Hardware type} & 
  \begin{itemize}[noitemsep, topsep=0pt]
  \item Mechatronic
  \item Robotic
  \end{itemize}\\
 \textbf{Open source license} & GNU GPL v3 \\
 \textbf{Cost of hardware} & $<$ \$1500\\
 \textbf{Source file repository} &\href{https://doi.org/10.17605/OSF.IO/K83X7}{https://doi.org/10.17605/OSF.IO/K83X7}\\
 \bottomrule
\end{tabular}
\end{center}
\label{tab:sec11}
\end{table}

\newpage

\section{Hardware in context}
Intelligent transport robots (ITRs) are becoming an integral part of our daily activities in recent years \cite{sell}, \cite{doi:10.1073/pnas.1805770115}, \cite{10.1007/978-3-030-50426-725}. Their application for day-to-day activities especially in industrial logistics \cite{logistics}, warehousing~\cite{FRAGAPANE2021405}, household tasks~\cite{household},~\cite{Jordi}, etc., provides not only a cleaner and safer work environment but also helps to reduce the high costs of production.  These robots offer the potential for a significant improvement in industrial safety~\cite{UNGER2018254}, productivity~\cite{Pawar2016ManufacturingRT} and general operational efficiency~\cite{Atkinson2019RoboticsAT}. However, several challenges such as the reduced capacity\cite{Pagliarini}, \cite{Atkinson2019RoboticsAT}, affordability \cite{Fragapane2022IncreasingFA}, and the difficulties in  modifying the inbuilt hardware and firmware still remain \cite{opentorque}.

Although significant effort has been undertaken by the scientific community in recent years to develop a standardised low-cost mobile platform, there is no open-source system that fulfils the high industrial requirements. Most available open-source, low-cost platforms are still limited in their functions and features, i.e., they are commercially not available, do not match the required payloads for logistic tasks, or cannot be adapted. For example, while~\cite{Hui2010},~\cite{BETANCURVASQUEZ2021e00217}, and~\cite{jo2022smartmbot}, are inexpensive, they cannot be used in day-to-day tasks that require transporting materials of high loads of 5kg or more.

On the other hand, many commercially available industrial platforms exist that feature high payloads, see Table~\ref{tab:3} for an overview. Unfortunately, they are expensive, closed-source, and platform-specific with inbuilt hardware, and firmware that may be difficult to modify~\cite{jo2022smartmbot}. This restrains many users' ability to explore multiple customisation or reconfiguration options to accelerate the development of intelligent systems.

Consequently, there is a crucial need to have low-cost robots with comparable features that can easily be scaled to adapt to any useful purpose. To this end, we propose $ROMR$, a modular, open-source and low-cost alternative for general-purpose applications including research, navigation \cite{agile-giss-3-55-2022}, and logistics. 

\begin{figure}[H]
   \centering
    \subfigure[]{\includegraphics[scale=0.32]{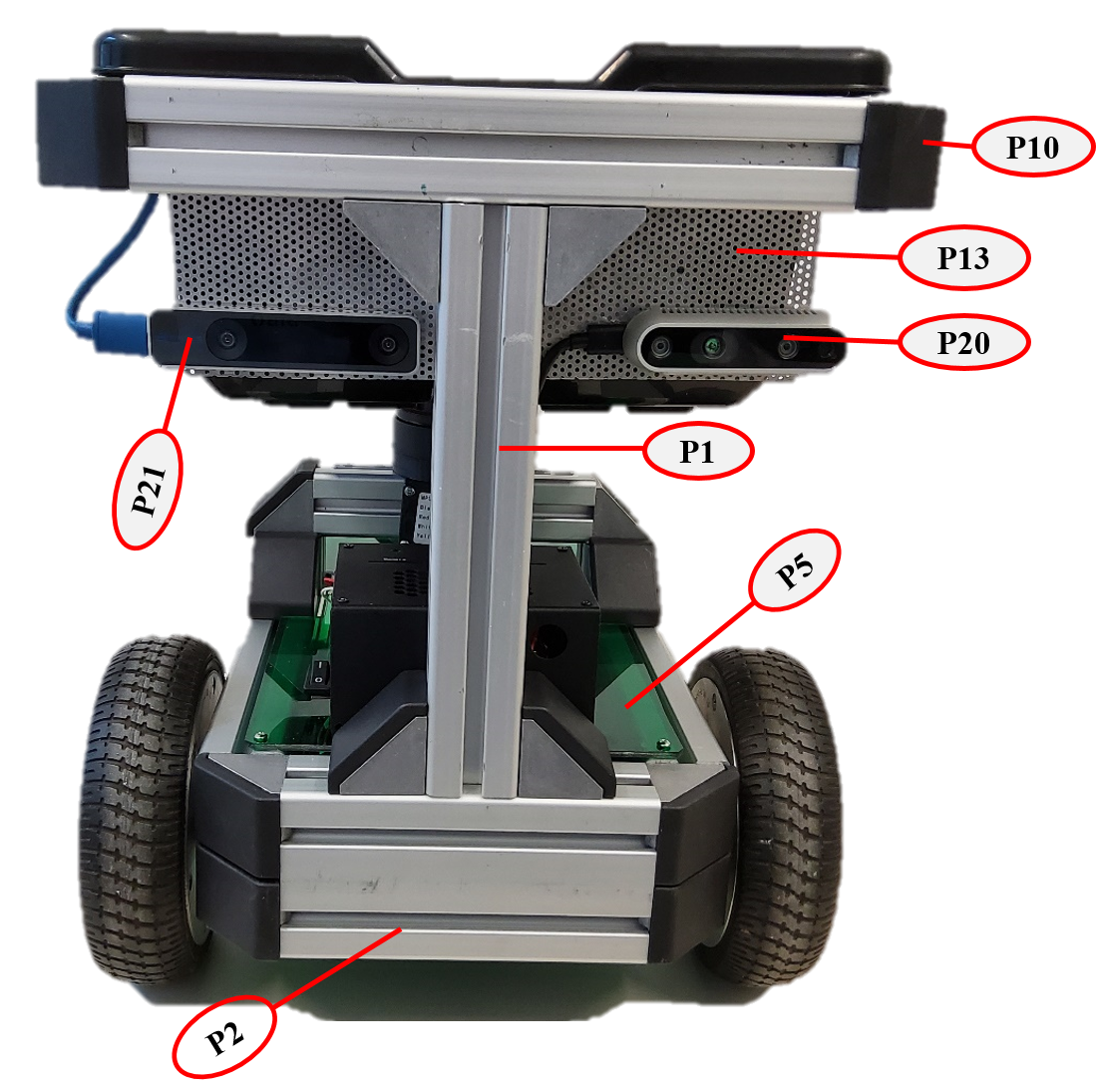}} 
    \subfigure[]{\includegraphics[scale=0.32]{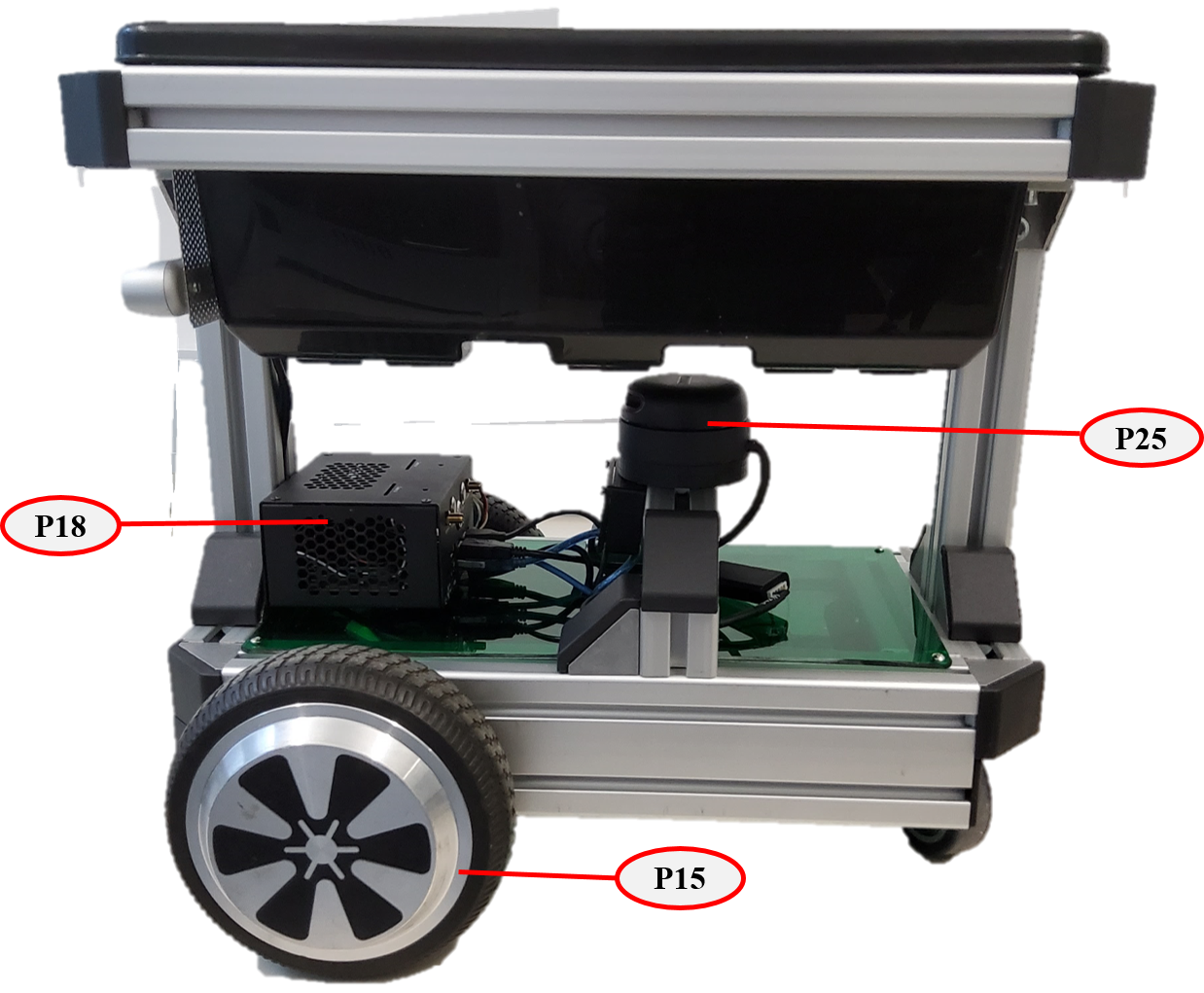}}
    \subfigure[]{\includegraphics[scale=0.30]{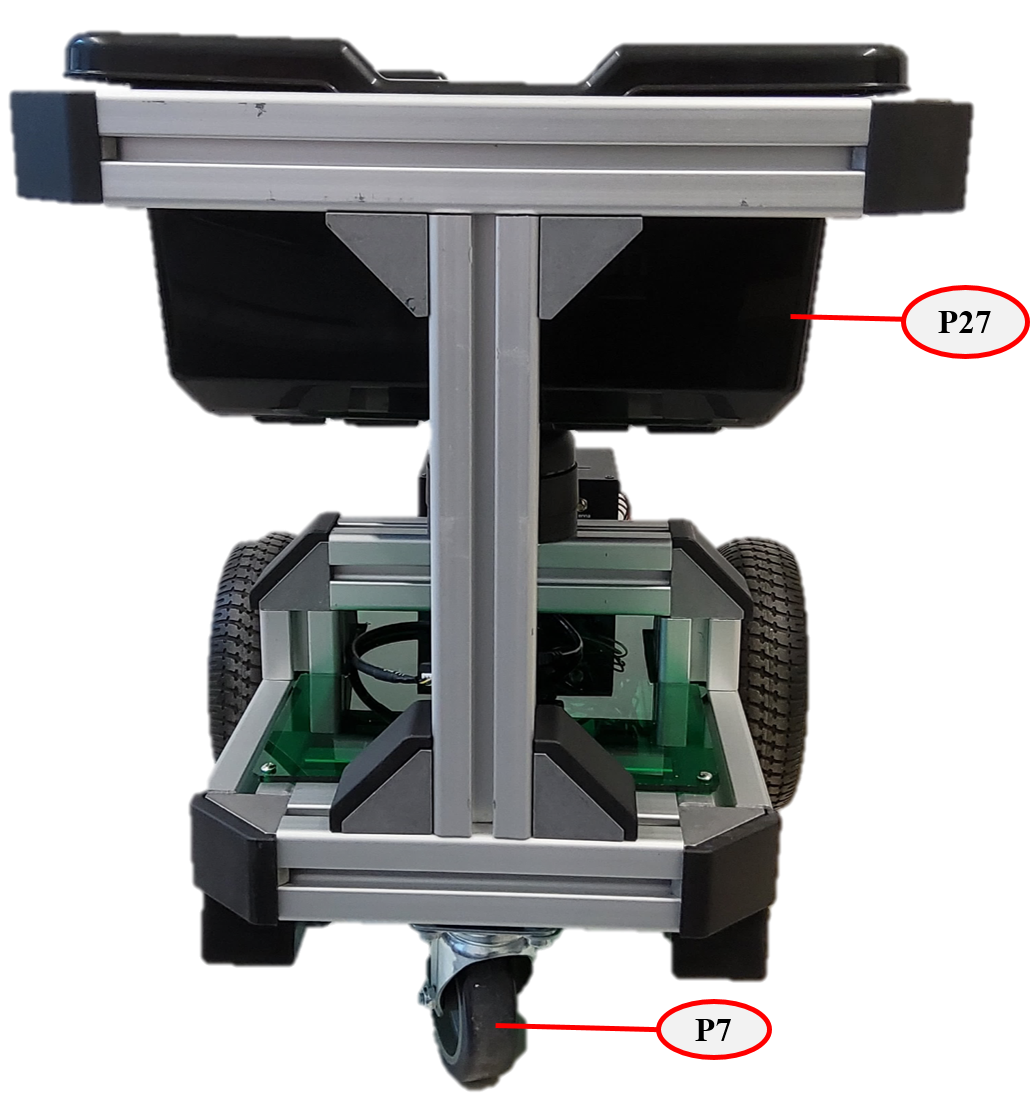}} 
    \caption{$ROMR$ is built from consumer hoverboard wheels with high torque brushless direct current (BLDC) motors. It utilises Arduino Mega Rev3, an Nvidia jetson Nano, and the components described in Tables \ref{tab:5} and \ref{tab:6}. (a) front view (b) side view (c) back view.}
   \label{fig:1}
\end{figure}

$ROMR$ is fully compatible with ROS, has a maximum payload of 90kg and costs less than \$1500. It features several lidar sensor technologies for potential application for perception \cite{Kolski2007}, simultaneous localisation and mapping \cite{slam}, \cite{thrunsabastian}, deep learning tasks \cite{shabbir2018survey}, and many more. Figures \ref{fig:1} and \ref{fig:wiring} present the pictorial view and the cyber-physical anatomy of $ROMR$ respectively.

\subsection{Related hardware platforms}
In Table \ref{tab:3}, we present a comparative evaluation with $ROMR$, similar hardware developed in recent years for research, navigation and logistics applications. Our comparison focused on some key features that define the open-sources, robustness and versatility of the robot design, e.g., ease of reconfiguration or modification, cost performance, load carrying capacity, and full compatibility with the ROS \cite{ros}.

In Table \ref{tab:3}, the ROS feature indicates whether the robot is fully compatible with ROS or not. Custom determines whether the platform satisfies easy modification of its design and integration of additional components. OpenS determines whether the hardware (electronic circuits, design files, etc.) and software (source codes, ROS packages, etc.) are fully open-source and maintained by the open-source community such that external hobbyists can replicate the same design without the need to contact the developer. Finally, the cost feature determines the affordability of the system. As shown in Table \ref{tab:3}, the majority of the robots are not open-source and are expensive. This limits wide application, e.g., in research, navigation and logistics. Our $ROMR$ has been developed as a low-cost open-source alternative. The approximate cost to redevelop it currently stands at less than \$1,500.00. Recently, $ROMR$ is been used for both B.Sc. and M.Sc. projects in our laboratory.

\arrayrulecolor{black}
\begin{table}[H]
\centering
\caption{Comparison of existing mobile platforms to our $ROMR$ development. }
\begin{center}
\begin{tabular}{p{0.2\linewidth} p{0.06\linewidth} p{0.15\linewidth} p{0.08\linewidth} p{0.1\linewidth} p{0.08\linewidth}}
  \toprule
    {\textbf{Robot Name}} & {\textbf{ROS}} & {\textbf{Payload ($kg$)}} & {\textbf{Custom}} & {\textbf{Cost $(k\$)$}} & {\textbf{OpenS}}\\
     \midrule
    RMP Lite 220 & \checkmark & 50 & \checkmark & 2.99 & x\\
    Ackerman Pro Smart & \checkmark & 22 & \checkmark & 3.99 & x\\
    Nvidia Carter & \checkmark & ~50 & \checkmark & 10.00 & \checkmark\\
    Panther UGV & \checkmark & 80 & \checkmark & 15.90 & x\\
    Tiago base & \checkmark & 100 & \checkmark & 11.50 & x\\
   Clearpath TurtleBot 4  & \checkmark & 9 & \checkmark & 1.90 & \checkmark \\
    Summit XL & \checkmark & 65 & \checkmark  & 11.50 & x\\
    MIR100 & \checkmark & 100 & \checkmark & 24.00 & x\\
    AgileX Scout 2.0 & \checkmark & 50 & \checkmark & 12.96 & x\\
    Jackal J100 & \checkmark & 20 & \checkmark & 18.21 & x\\
    4WD eXplorer & \checkmark & 90 & \checkmark & 15.70 & x\\
    ROSbot 2.0 & \checkmark & 10 & \checkmark & 2.34 & x\\
    $ROMR$ & \checkmark & 90 & \checkmark & 1.50 & \checkmark\\
    \bottomrule
\end{tabular}
\end{center}
\label{tab:3}
\end{table}

\section{The ROMR description}
$ROMR$ is compactly and robustly designed to ensure stability, ease of integration of additional components, and low cost of reproducing the system. In this section, we describe the robot's hardware and software platforms. Afterwards, we describe the technical specifications and tools, as well as the detailed architecture of the control unit.

\subsection{Hardware description}\label{sec21}
We leveraged off-the-shelf (OTS) electronics that are commercially available online, additive manufacturing technologies (3D printing), and aluminium profiles for the robot's structural design. The main reason for using aluminium profiles is to achieve a lightweight structure that can hold the hardware and associated electronics of the robot without increasing its overall weight. At the same time, the profiles could resist load stress and damage during everyday use. The profile bars with slots were also used to enclose all the electronics and power subsystems at the base of the robot for proper weight distribution. This increases the flexibility to connect any additional hardware component to it. 

$ROMR$ is equipped with an Arduino Mega Rev 3, an Nvidia Jetson Nano board, an Odrive 56V V3.6 brushless direct current (DC) motor controller, and two 350W hoverboard brushless motors with five inbuilt hall sensors. The Arduino board is responsible for low-level tasks, such as gesture-based control of the robot using an inertial measurement unit (IMU) sensor and teleoperation from remote-controlled (RC) devices. In contrast, the Jetson Nano board handles high-level processing tasks such as deep learning, SLAM, and ROS navigation tasks with the RGB-D cameras and LiDARs, etc. The two boards communicate with each other through the serial UART interface.

Furthermore, $ROMR$ is powered by rechargeable lithium-ion batteries (36V 4400mAh), which are affordable, inexpensive to maintain, and eco-friendly (i.e., they do not contain heavy metals such as lead or cadmium which are harmful to the environment and human health).  Additionally, the robot is endowed with an Intel Realsense D435i RGB-D camera for visual perception and depth sensing, and an Intel Realsense T265 for localization and tracking. $ROMR$ is also equipped with a 9-axis MPU 9250 IMU sensor for tracking and localisation. The IMU sensor is used also for gesture-based teleoperation which allows a non-robotics expert to intuitively control the robot using hand gestures (see Subsection \ref{sec733} for more details). For 2D mapping, we used an RPlidar A2 M8 with a 360-degree field of view (FOV). This lidar has a maximum range distance of $16 m$ and operates at a frequency of $10 Hz$. The lidar sensor allowed us to create a 2D occupancy-grid map of the environment, which was then used to support the $ROMR$ navigation, localization, and obstacle detection within the environment.

We supported $ROMR$ with a small caster wheel in addition to its drive wheels to increase stability, manoeuvrability, and proper weight distribution. Although it is possible to use a bigger or two caster wheels, however, we chose a smaller caster wheel to make it easier to navigate around tight spaces and obstacles and provide better stability and balance for the robot, especially if it has to change direction quickly or make sudden turns. Furthermore, by using a smaller caster wheel, the weight of the robot could be distributed more evenly, which can help to prevent tipping or loss of balance. Note, it is recommended to add a piece of rubber between the caster wheel and the $ROMR$'s base frame to compensate for hyperstaticity.

\subsection{Software description}
The robot's main software is based primarily on the ROS framework, which runs on both Ubuntu 18.04 (ROS Melodic version) and Ubuntu 20.04 (ROS Neotic version). The ROS framework provides a set of tools, libraries, and conventions for building the robot system.

The software subsystems include the Arduino sketches, ODrive calibration programs, the ROS workspace containing the $ROMR$ universal robot description format (URDF) files for a Gazebo simulation, the Gazebo plugins files, the $ROMR$ meshes, the launch files, the joint and rviz configuration files, and the RPlidar packages.  The files are listed in Table \ref{tab:10} and are published using the open-source license GNU GPL V3, which allows the community to reproduce, modify, redistribute, and republish them.

\subsection{Control unit description}
The control unit includes multiple options for controlling the robot. In addition to the existing hardware described in Subsection \ref{sec21}, the control unit includes RC receivers and transmitters, an Android device running a ROS-mobile app, an IMU sensor, an nRF24L01+ module, and an additional Arduino board. The RC receivers and transmitters are used to provide manual control of the robot. The Android device running the ROS-mobile app serves as an alternative control interface for the robot. The nRF24L01+ module provides wireless communication between the control unit and the robot. The additional Arduino board is used to interface with the nRF24L01+ module and IMU sensor to handle wireless communication with the robot in case of gesture-based teleoperation. Subsections \ref{sec731}, \ref{sec732}, \ref{sec733}, and Figures \ref{fig:rc_test}, \ref{fig:ros-mobile} and \ref{fig:gesture} provide insights into the architecture of the control unit.

\subsection{Technical specifications and features}
The technical specifications of $ROMR$ are summarized in Table \ref{tab:4}.
\begin{table}[h]
\caption{$ROMR$ technical specifications.  $L  \rightarrow$ Length, $W  \rightarrow$ Width, $H  \rightarrow$ Height, $\phi \rightarrow$ Wheel diameter.}
\begin{center}
\begin{tabular}{p{0.25\linewidth} p{0.4\linewidth}}
  \toprule
    {\textbf{Parameters}} & {\textbf{Technical specifications}}\\
     \midrule
    Robot dimensions & $ L * W * H  = 0.46\;m * 0.34\;m * 0.43\;m $\\
    Wheel dimensions & Drive wheel $(\phi = 0.165\;m)$; caster wheel $(\phi = 0.075\;m) $\\
    Inter-wheel distance & $0.29\;m$ \\
    Robot weight &  $17.1\;kg$ \\
    Max. payload & $90\; kg$ \\
    Max. speed & Up to $3.33\;m/s$ \\
    Max. stable speed & $<$ $2.5\;m/s$ \\
    Battery capacity & $36\;V$, $4400\;mAh$ \\
    Motor type  & BLDC with 15 pole pairs (350W x 2)  \\
    Ground clearance & $0.065\;m$  \\
    Operation environment & Indoor and outdoor  \\
   Run time (full charge) & Approximately 8 hours with the robot weight ($ 17.1 kg $) only \\
    \bottomrule
\end{tabular}
\end{center}
\label{tab:4}
\end{table}
\arrayrulecolor{black}
\begin{table}[h]
\caption{Overview of the $ROMR$ hardware and software tools.}
\begin{center}
\begin{tabular}{p{0.25\linewidth} p{0.58\linewidth}}
  \toprule
    {\textbf{Features}} & {\textbf{Tools}}\\
     \midrule
    Actuation  & ODrive 56V V3.6 brushless DC motor controller, Nvidia Jetson Nano, and Arduino Mega Rev 3 \\
    Sensing \& feedback  & RPlidar A2, IMU, Depth cameras (Intel Realsense D435i \& T265) \\
    Operating system  & Ubuntu 20.04 (ROS Neotic) or Ubuntu 18.04 (ROS Melodic)\\
    Communication & ROS architecture (ROS C/C++ \& ROS Python libraries), WiFi 802.11n, USB \& Ethernet (for debugging)\\
    Navigation \& drive interfaces & ROS navigation stack, position \& joint trajectory controller, joystick, rqt-plugin, ROS-Mobile (Android devices), hand gesture, web-based GUI\\
    SLAM  & Hector-SLAM, Cartographer, Gmapping, RTAB-Map, etc \\
    Simulation \& visualisation & Gazebo, Rviz, MATLAB\\
    \bottomrule
\end{tabular}
\end{center}
\label{tab:2}
\end{table} 
Furthermore, in Table \ref{tab:2}, we present the tools and key features of $ROMR$ in line with other robots with comparable specifications. Initially, when envisioning the design, one of our core goals was to develop a low-cost scalable platform that robotic developers and the open-source community could easily adapt, to foster research in mobile navigation. For this reason, we tailored our design considerations based on this goal such that the $ROMR$ should be:
\begin{itemize}
    \item Modular - To offer the users the opportunity to easily integrate additional parts or units and reconfigure them to suit their needs.
    \item Portable and simple - To ensure that minimal and off-the-shelf hardware components could be used for its construction and replication. This would allow users not to worry about purchasing costly components and instead focus on the system's functional design.
    \item Low-cost and open-source - To ensure affordability and commercialisation of the system so that users can leverage the $ROMR$ framework in any form to develop novel and trivial robotic applications.
    \item Versatile and suitable - To ensure that it takes minimal time to be re-programmed for any useful purpose, whether navigation, logistics etc as well as adapt to new processes and changes in the environment.
    \item Unique - To enable hobbyists and users to learn new tools, techniques and methods useful to accelerate the development of intelligent systems.
\end{itemize}

\section{Design files summary}
The $ROMR$ design files are categorised into three different units, (a) the mechanical unit, (b) the software unit, and (c) the power, sensors and electronics unit. Each of these units is briefly described in Tables \ref{tab:5}, \ref{tab:6} and \ref{tab:10}. Table \ref{tab:5} describes the additively manufactured part (3D printed) and the 3D CAD models of the aluminium profiles and their accessories. The CAD files were used to generate the universal robot description format (URDF) \cite{urdf} description of the $ROMR$ in order to simulate it using the ROS framework \cite{ros}. These parts were designed using the Solid Edge CAD tool. Table \ref{tab:6} lists all off-the-shelf (OTS) electronics components. Table \ref{tab:10} contains the software files. All design and construction files can be downloaded at our repository: \url{https://doi.org/10.17605/OSF.IO/K83X7}.

\subsection{Mechanical unit}
The mechanical unit includes the additively manufactured parts, the 3D CAD models of the aluminium profiles, and their accessories as summarised in Table \ref{tab:5}. The labels $P_n$ with $n = 1, 2, . . .$, refers to the individual parts.
\arrayrulecolor{black}
\begin{table}[ht]
\caption{Summary of the $ROMR$ mechanical structure unit.}
\begin{center}
\begin{tabular}{p{1.2cm} p{4.2cm} p{2.5cm} p{3.0cm} p{4.0cm}}
  \toprule
   {\textbf{Des.}} & {\textbf{Description}} & {\textbf{File type}} & {\textbf{Open S. license}} & {\textbf{File location}}\\
     \midrule
  P1 & Alum. 40x40mm slot 8 & .stp & GNU GPL v3  & \url{https://osf.io/qpb2v} \\
  P2 & Alum. 40x80mm slot 8 & .stp & GNU GPL v3 & \href{https://osf.io/zmqe6}{https://osf.io/zmqe6} \\
  P3 &  Corner bracket I-type & .stp & GNU GPL v3 & \href{https://osf.io/63jwa}{https://osf.io/63jwa}\\
  P4 &  Mounting bracket I-type & .stp & GNU GPL v3 & \url{https://osf.io/ncbu9}\\
  P5 &  Bottom cover & .stp & GNU GPL v3 & \url{https://osf.io/ae9um}\\
  P6 &  Top cover & .stp & GNU GPL v3 & \url{https://osf.io/9v5j4}\\
  P7 &  Swivel caster  & .stp & GNU GPL v3 & \url{https://osf.io/g7pxb} \\
  P8 & Screw/bolt & .stp & GNU GPL v3 & \url{https://osf.io/jcefn} \\
  P9 &  Corner bracket cover cap & .stp & GNU GPL v3 & \url{https://osf.io/sfgb9}\\
  P10 & $ROMR$ full assemble & .asm & GNU GPL v3 & \url{https://osf.io/jybzf} \\
  P11 &  0.42x0.32x0.15m box & .stp & GNU GPL v3 & \url{https://osf.io/5v3ra}\\
  P12 &  Lock nut & .stp & GNU GPL v3 & \url{https://osf.io/dk3ha}\\
  P13 &  Front hole plate & .stp  & GNU GPL v3 & \url{https://osf.io/yhuw9} \\
   P14 & RPLidar base holder & .stl (3D print) & GNU GPL v3 & \url{https://osf.io/envdh} \\
  P15 &  Drive wheel & .stp & GNU GPL v3 & \url{https://osf.io/vwztk} \\
 \bottomrule
\end{tabular}
\end{center}
\label{tab:5}
\end{table}
\begin{itemize}
    \item $P1 - P4$, $P7 - P9$ and $P12$ are aluminium profiles and accessories used for the robot's chassis construction. 
    \item $P5$ and $P6$ are used to cover the base of the robot, where the electronic components are placed.
    \item $P10$ is the full CAD assembly of the $ROMR$.
    \item $P11$ and $P13$ are used for material carriage and for mounting the RGB-D cameras respectively.
    \item $P14$ is 3D-printed to attach the RPlidar sensor.
   \item $P15$ is the $ROMR$ drive wheel from a consumer hoverboard scooter.
\end{itemize}

\subsection{Power, sensors, and electronics units}\label{sec32x}
Table \ref{tab:6} shows a summary of the used electronics, sensors and control devices, as well as the power sources. $P16 - P27$ are off-the-shelf (OTS) components from different vendors. All vendors are listed in Table \ref{tab:7}.
\arrayrulecolor{black}
\begin{table}[h]
\caption{OTS electronics, sensors and control devices.}
\begin{center}
\begin{tabular}{p{1.2cm} p{6cm} p{2.2cm} p{1.2cm} p{3.2cm}}
  \toprule
    {\textbf{Des.}} & {\textbf{Description}} & {\textbf{File type}} & {\textbf{Qty}} & {\textbf{File location}}\\
     \midrule
   P16 & Nvidia Jetson Nano B01 64GB & png & 1 & \href{https://osf.io/72xtm}{https://osf.io/72xtm} \\
  P17 & Arduino Mega Rev 3 & png & 1 & \href{https://osf.io/d3qmj}{https://osf.io/d3qmj} \\
   P18 & ODrive V3.6 56V & png & 1 & \href{https://osf.io/jghnv}{https://osf.io/jghnv} \\
   P19 & IMU (Invensense MPU-9250 9DOF) & png & 1 & \href{https://osf.io/k2h35}{https://osf.io/k2h35}\\
  P20 & Intel Realsense D435i camera & png & 1 & \href{https://osf.io/xu368}{https://osf.io/xu368}\\
  P21 & Intel Realsense T265 camera & png & 1 & \href{https://osf.io/6cj5r}{https://osf.io/6cj5r} \\
  P22 & nRF24L01+PA+LNA module & png & 1 & \href{https://osf.io/43kd6}{https://osf.io/43kd6} \\
  P23 & Turnigy 2.4GHz 9X 8-Channel V2 transmitter \& receiver & png & 1 & \href{https://osf.io/3wu2x}{https://osf.io/3wu2x} \\
  P24 & 125mm \& 225mm M2M, M2F, F2F GPIO wires & png & 24 & \href{https://osf.io/kv982}{https://osf.io/kv982} \\
  P25 & RPLidar A2 M8 & png & 1 & \href{https://osf.io/pej62}{https://osf.io/pej62} \\
  P26 & 36V Lithium Ion battery 4400mAh & png & 1 & \href{https://osf.io/umskh}{https://osf.io/umskh} \\
  P27 & Power bank 2400mA & png & 1 & \href{https://osf.io/z98gv}{https://osf.io/z98gv} \\
     \bottomrule
\end{tabular}
\end{center}
\label{tab:6}
\end{table}
\begin{itemize}
\item $P16$ is the main brain of the robot. It handles the high-level control task required to run all the sensing, perception, planning and control modules.
\item $P17$ is one of the most successful open-source platforms with relatively easy-to-use free libraries compared to other open-source micro-computers. It is used for the low-level control, to send the control command to $P18$ which in turn drives and steers the $P15$. Furthermore, it was used to handle communication between $P16$ and $P18$, as well as for any future compatible devices which support ROS serial communication e.g., Raspberry Pi, STM32, etc.
\item $P18$ is a high-performance, open-source brushless direct current (BLDC) motor driver from ODrive robotics \cite{odrive}. It regulates all computations required to drive the two inbuilt hoverboard brushless DC motors with five hall-effect sensors. The sensors are used for the motor position feedback. Note that in the case of the $P18$ end-of-life (EOL), the Odrive Pro found at \url{https://odriverobotics.com/shop/odrive-pro} can be used as the motor driver replacement since it provides even more advanced features than the one used in this work.
\item $P19$ is a 9-axis inertial measurement unit (IMU) sensor specifically used for tracking, localization and gesture-based control tasks.
\item $P20$ and $P21$ are 3D vision cameras for visual perception, depth sensing, tracking, localization and mapping tasks. Both cameras generate 3D image data of the task environment, process the data and then publish it to the appropriate topic in the ROS network.
\item $P22$ and $P23$ are communication devices for wireless control of the $ROMR$.
\item $P24$ are general-purpose input-out (GPIO) connection wires.
\item $P25$ is a 2D lidar with a 360-degree field of view (FOV), a maximum range distance of 16m, operating at a frequency of 10Hz. It is used for generating a 2D occupancy grid map of the robot's operational environment and for collision detection and avoidance.
\item $P26$ and $P27$ are the system's power sources for the motors, the sensors, and the control boards.
\end{itemize}

\arrayrulecolor{black}
\begin{table}[ht]
\caption{Software files.}
\begin{center}
\begin{tabular}{p{0.05\linewidth} p{0.15\linewidth} p{.20\linewidth} p{0.10\linewidth} p{0.15\linewidth} p{0.18\linewidth}}
  \toprule
    {\textbf{Des. Name}} & {\textbf{File Name}} & {\textbf{Description}} & {\textbf{Type}} & {\textbf{Open source license}} & {\textbf{File location}}\\
     \midrule
    S1 & sketches.zip & Folder containing the Arduino sketches & Arduino sketches & GNU GPL v3 & \url{https://osf.io/r5cgp} \\
    S2 & romr\_robot.zip & Folder containing the $ROMR$ ROS files & ROS files & GNU GPL v3 & \url{https://osf.io/e4syc} \\
        \bottomrule
\end{tabular}
\end{center}
\label{tab:10}
\end{table}
\begin{itemize}
    \item S1 and S2 are the folders containing the Arduino control programs and the $ROMR$ ROS files respectively.
\end{itemize}

\section{Bill of materials (BOM) summary}\label{sec4}
Table \ref{tab:7} provides a summary of the bill of materials, which includes the lidar, the depth cameras, the Turnigy 2.4GHz 9X 8-Channel V2 transmitter \& receiver, the 22nF capacitors, and the MPU-9250 sensor, that were sourced in the laboratory. The BOM reflects only the component prices and does not include labour costs (purchasing, manufacturing, marketing, ...).

\arrayrulecolor{black}
\begin{table}[ht]
\caption{BOM for building $ROMR$, and the respective links of where they were purchased. The BOM reflects only the component prices and does not include labour costs (purchasing, manufacturing, marketing, ...).}
\begin{center}
\begin{tabular}{p{0.25\linewidth} p{0.06\linewidth} p{1.80cm} p{0.10\linewidth} p{0.22\linewidth} p{0.08\linewidth}}
  \toprule
    {\textbf{Designator}} & {\textbf{Qty}} & {\textbf{Unit cost (\euro)}} & {\textbf{Total cost (\euro)}} & {\textbf{Source of material}} & {\textbf{Material type}}\\
     \midrule
    Profile 40x40L I-type slot 8 (1.98m long) & 1 & 26.49 & 26.49 & \href{https://www.motedis.at/shop/Slot-profiles/Profile-40x40L-I-Type-slot-8::999999.html}{www.motedis.at} & Other\\
    Profile 40x80L I-type slot 8 (1.1m long) & 1 & 25.64 & 25.64 & \href{https://www.motedis.at/shop/Slot-profiles/Profile-40x80L-I-type-slot-8::99999456.html}{www.motedis.at} & Other\\
    Corner bracket I-type & 20 & 0.51 & 10.20 & \href{https://www.motedis.com/en/Bracket-40-I-Type-slot-8}{www.motedis.at} & Other\\
     Mounting bracket I-type & 40 & 0.18 & 7.20 & \href{https://www.motedis.com/en/T-nut-guided-I-type-slot-8-M8}{www.motedis.at} & Other\\
      Bottom \& Top cover (50 x 50 cm) & 1 & 6.99 & 6.99 & \href{https://www.obi.at/kunststoffbedachung/hobbyglas-0-2-cm-transparent-50-cm-x-50-cm/p/7688096}{https://www.obi.at/} & Other\\
       Screw/bolt & 100 & 0.13 & 12.64 & \href{https://www.motedis.com/en/Screw-DIN-912}{www.motedis.at} & Other\\
        Corner bracket cover cap & 20 & 0.18 & 3.60 & \href{https://www.motedis.com/en/Covers-Bracket-covers-covers-cap}{www.motedis.at} & Other\\
   Swivel caster  & 1 & 4.51 & 4.51 & \href{https://www.motedis.at/shop/Mechanical-Basics/Wheels-and-rollers/Roller-75-plate-fitting-with-brake::99999395.html}{www.motedis.at} & Other \\
    Hoverboard brushless DC motor wheels & 2 & 24.50 & 49.00 & \href{https://www.voltes.nl/en/hoverboard-wheel-6-5-inch.html}{www.voltes.nl} & Other \\
    Nvidia Jetson Nano B01 64GB  & 1 & 260.22 & 260.22 & \href{https://www.amazon.de/-/en/MakerFun-Developer-Computer-AC8265-WLAN-Development/dp/B08Y68WVVG/ref=sr_1_4?crid=ZJK2SIKQ0PDH&dchild=1&keywords=jetson+nano&qid=1634660329&qsid=257-7183770-6672511&sprefix=jets\%2Caps\%2C202&sr=8-4&sres=B08M5J1WM2\%2CB08Y68WVVG\%2CB081CN3VB1\%2CB08J157LHH\%2CB07PZHBDKT\%2CB08BFVY5YP\%2CB085W1RXMJ\%2CB07QWLMR24\%2CB084P23M3R\%2CB084DSDDLT\%2CB088H3Q2Q6\%2CB098J4JMLG\%2CB083FP9JRY\%2CB095S3VP9W\%2CB07SJLWWT3\%2CB08MTLQCLJ}{www.amazon.de} & Other \\
    Arduino Mega Rev 3  & 1 & 32.78 & 32.78 & \href{https://www.reichelt.at/at/de/arduino-mega-2560-atmega-2560-usb-arduino-mega-p119696.html?PROVID=2807&gclid=Cj0KCQjwpeaYBhDXARIsAEzItbFMGkM8d1ArrDtVmPL-hsIv3c_eOmU02YTgx57oYCcr_YW_23JGox4aAr7iEALw_wcB}{www.amazon.de} & Other\\
    ODrive V3.6 56V & 1 & 249.00 & 249.00 & \href{https://odriverobotics.com/shop/odrive-v36}{www.odriverobotics.com} & Other\\
    IMU (MPU-9250)  & 1 & 15.92 & 15.92 & \href{https://www.distrelec.at/de/grove-imu-10dof-v2-seeed-studio-101020252/p/30109497}{www.distrelec.at} 
    & Other\\
    nRF24L01+PA+LNA module & 1 & 9.99 & 9.99 & \href{https://www.amazon.de/DollaTek-NRF24L01-Funk-Transceiver-Modul-antistatischem-kompatibel/dp/B07PQPFTWZ/ref=asc_df_B07PQPFTWZ/?tag=googshopde-21&linkCode=df0&hvadid=341227547250&hvpos=&hvnetw=g&hvrand=9299400644923912530&hvpone=&hvptwo=&hvqmt=&hvdev=c&hvdvcmdl=&hvlocint=&hvlocphy=1000912&hvtargid=pla-1189460927948&psc=1&th=1&psc=1&tag=&ref=&adgrpid=66631953017&hvpone=&hvptwo=&hvadid=341227547250&hvpos=&hvnetw=g&hvrand=9299400644923912530&hvqmt=&hvdev=c&hvdvcmdl=&hvlocint=&hvlocphy=1000912&hvtargid=pla-1189460927948}{www.amazon.de} & Other\\
    Turnigy 2.4GHz 9X 8-Channel V2 transmitter \& receiver & 1 & 69.99 & 69.99 & \href{https://hobbyking.com/en_us/turnigy-9x-9ch-transmitter-w-module-ia8-receiver-mode-1-afdhs-2a-system.html}{www.hobbyking.com} 
    & Other\\
    125mm \& 225mm M2M, M2F, F2F GPIO wires & 1 & 3.99 & 3.99 & \href{https://www.amazon.de/AZDelivery-Jumper-Arduino-Raspberry-Breadboard/dp/B07KYHBVR7/ref=asc_df_B07KYHBVR7/?tag=&linkCode=df0&hvadid=308872732263&hvpos=&hvnetw=g&hvrand=13882554571830420241&hvpone=&hvptwo=&hvqmt=&hvdev=c&hvdvcmdl=&hvlocint=&hvlocphy=1000912&hvtargid=pla-702059949303&th=1&ref=&adgrpid=61255173493}{www.amazon.de} & Other\\
     36V Lithium-Ion battery 4400mAh & 1 & 33.63  & 33.63  & \href{https://de.aliexpress.com/item/1005003444525887.html?spm=a2g0o.productlist.main.73.7f965196QTEZuB&algo_pvid=c72cfca3-ed84-45b0-84d9-44e8b4ec3a56&algo_exp_id=c72cfca3-ed84-45b0-84d9-44e8b4ec3a56-36&pdp_ext_f=\%7B\%22sku_id\%22\%3A\%2212000026824709521\%22\%7D&pdp_npi=2\%40dis\%21EUR\%2168.64\%2133.63\%21\%21\%217.75\%21\%21\%40211bd3cb16636098813805814d0711\%2112000026824709521\%21sea&curPageLogUid=ce9R7S7HEBAd}{https://de.aliexpress.com} & Other\\
      22nF capacitors & 6 & 0.05  & 0.30  & \href{https://www.conrad.at/de/p/kemet-c315c223m1u5ta-keramik-kondensator-radial-bedrahtet-22-nf-50-v-20-l-x-b-x-h-3-81-x-2-54-x-3-14-mm-1-st-1420314.html}{https://www.conrad.at} & Other\\
        \bottomrule
\end{tabular}
\end{center}
\label{tab:7}
\end{table}

\section{Build instructions}\label{sec5}
Once the materials described in Table \ref{tab:5} and Table \ref{tab:6} are available, the subsequent task is to assemble them accordingly. To do that, the sequence of steps presented in this section should be followed. Some basic tools such as pliers, screwdrivers, a wire stripper, a 3D printer, a saw, a soldering iron and others are required for building the hardware.

\subsection{Hardware build instruction}
The building of the $ROMR$ chassis is done using the parts described in Table \ref{tab:5}. To assemble the chassis, we considered the system's compactness to maintain an acceptable mode of operation for a light and reliable system not compromising stability. The mechanical construction is done in several steps as follows:
\begin{enumerate}
 \item Top base assembly:  For this step, the parts required are $P1$, $P3$, and $P8$ as referenced in Table \ref{tab:5}.  First, cut out two each 0.38m, 0.26m and 0.24m lengths of $P1$. For each end of the piece, use a tap to cut thread and a clearance hole of a diameter of 0.008m (M8). A description of how to use a tap can be found here \href{https://www.wikihow.com/Use-a-Tap}{https://www.wikihow.com/Use-a-Tap}. Assemble the parts as illustrated in Figure \ref{fig:step1}, by following the direction of the arrows. The first block shows the components required, the middle block represents the exploded view with the interconnection of the components, and finally, the last block shows the outcome of the assembly.
 
 \begin{figure}[h]
    \centering
    \includegraphics[scale=0.55]{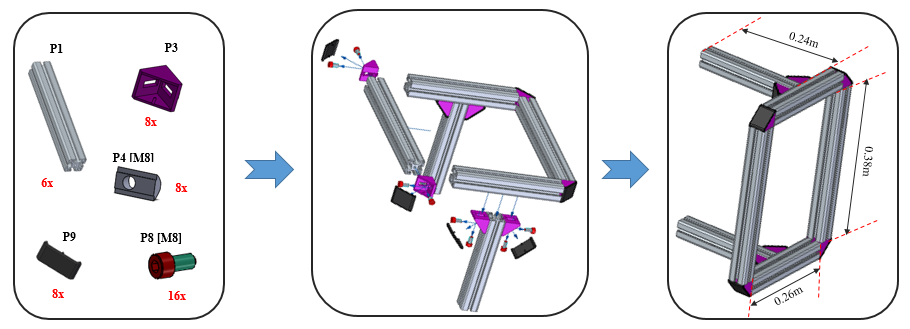}
    \caption{Top base chassis assembly.}
    \label{fig:step1}
\end{figure}

 \item Bottom and top chassis assembly: Parts required in this step are $P2$, $P3$, $P4$, $P8$, $P9$ and the result of step 1. Assemble the parts as illustrated in Figure \ref{fig:step2}.
 
  \begin{figure}[h]
    \centering
    \includegraphics[scale=0.42]{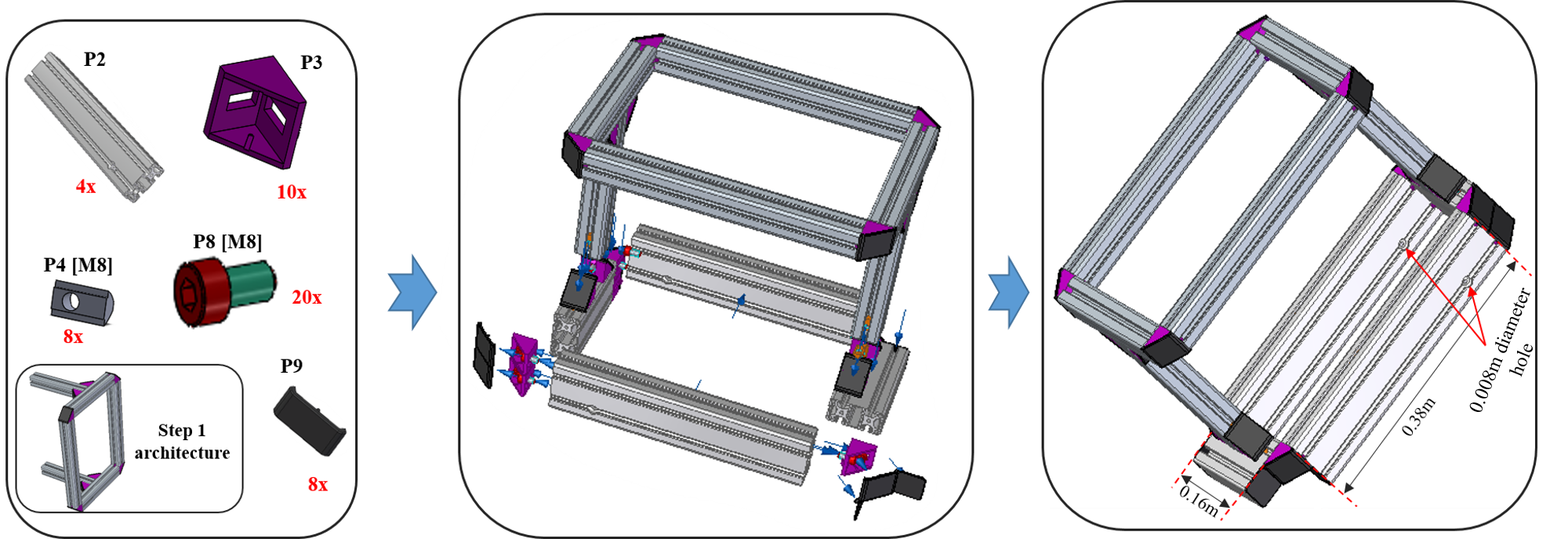}
    \caption{Assembling the top and bottom chassis.}
    \label{fig:step2}
\end{figure}

  \item Mounting of the caster and the drive wheels: Parts required are $P4$, $P7$, $P8$, $P15$ and the base from step 2. Note that $P8$ $(M8)$ and $P8$ $(M6)$ are required to mount $P15$ and $P7$ firmly to the base respectively. Assemble the parts as illustrated in Figure \ref{fig:step3}.
  
  \begin{figure}[h]
    \centering
    \includegraphics[scale=0.55]{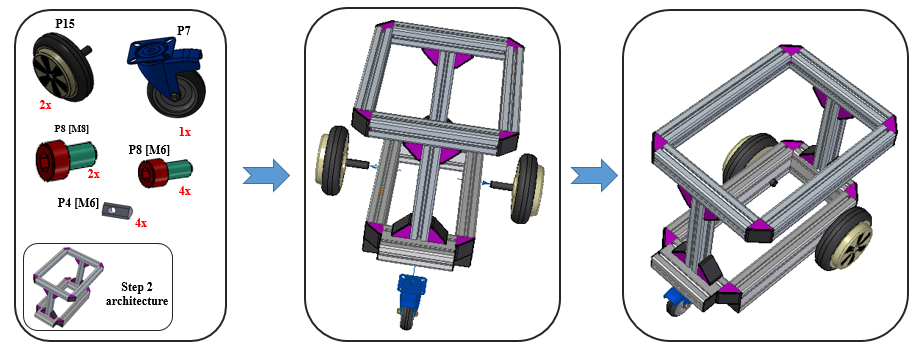}
    \caption{Mounting of the caster and the drive wheels to the base frame.}
    \label{fig:step3}
\end{figure}

  \item Mounting the bottom base plate: The base plate is rectangular green plastic meant to hold and protect all the electronics subsystems. Parts required are $P5$, $P8$ $(M3)$ and the resulting base from step 3. These parts are assembled as shown in Figure \ref{fig:step4}.
  
  \begin{figure}[h]
    \centering
    \includegraphics[scale=0.55]{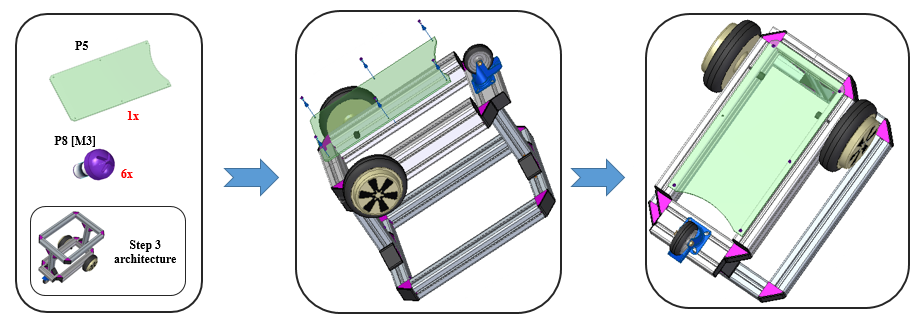}
    \caption{Mounting of the bottom plate to the base frame for attaching the electronics subsystems.}
    \label{fig:step4}
\end{figure}

  \item Mounting the internal electronics, the power subsystems to the bottom base plate and the RGB-D cameras: Parts required are $P13$, $P17$, $P18$, $P19$, $P20$, $P21$, $P22$, $P23$, $P26$, and the result from step 4. A couple of M3 screws (P8) are required for mounting the electronic components at the respective position. Figure \ref{fig:step5} illustrates the procedure.
  
  \begin{figure}[h]
    \centering
    \includegraphics[scale=0.55]{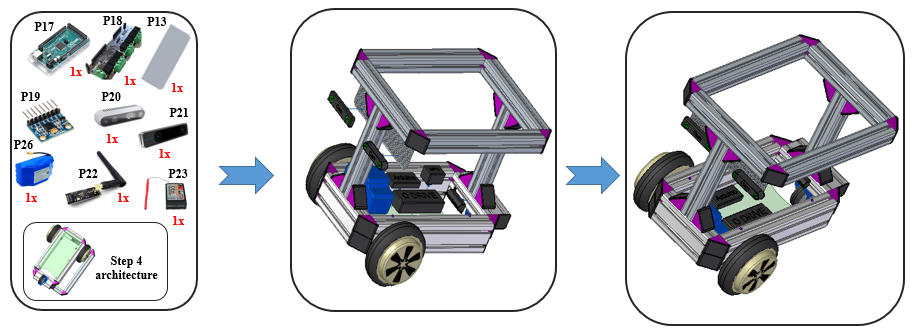}
    \caption{Mounting of the internal electronics and power subsystems to the bottom base plate.}
    \label{fig:step5}
\end{figure}

\item Mounting the top cover, external electronics, and final coupling: Parts required are $P1$, $P3$, $P4$, $P6$, $P8$ $(M3)$, $P8$ $(M8)$, $P9$, $P11$, $P16$, $P25$, and the resulting hardware from step 5. Figure \ref{fig:step6} illustrates the procedure.

\begin{figure}[h]
    \centering
    \includegraphics[scale=0.55]{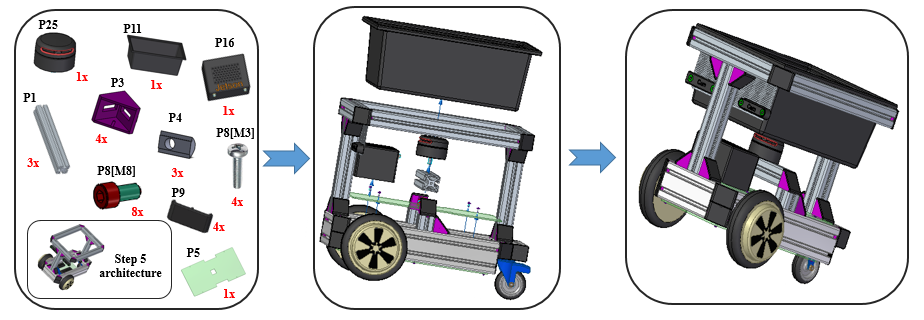}
    \caption{Mounting of the external electronics and the final coupling.}
    \label{fig:step6}
\end{figure}
\end{enumerate}

\subsection{General connection and wiring instruction}\label{sec52}
\begin{figure}
    \centering
    \includegraphics[scale=0.55]{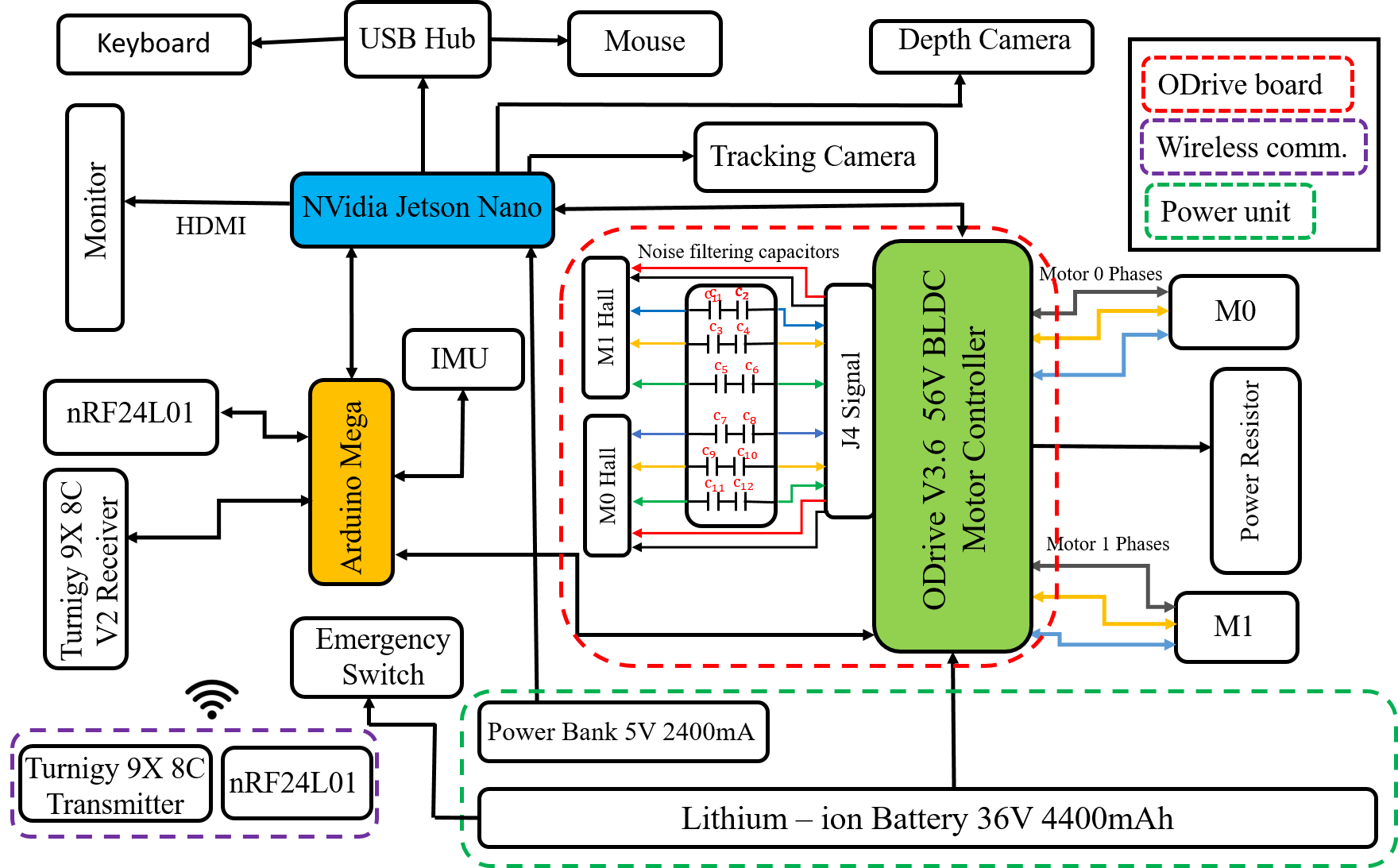}
    \caption{The $ROMR$ hardware architecture is based on an ODrive board (light green) to actuate the motors, an Nvidia Jetson Nano (light blue) as a computing interface for high-level tasks, an Arduino Mega Rev3 (orange) as low-level computing interface.}
    \label{fig:wiring}
\end{figure}

The electronics, vision and sensor subsystem of $ROMR$ are composed of several components interconnected by energy links and bidirectional or unidirectional information links as shown in Figure \ref{fig:wiring}. 
Each link either sends or receives information from the interconnected components. The instruction below shows how the system wiring was done.
\begin{enumerate}
      \item ODrive BLDC controller (P18): The ODrive controller has to be wired to the motors as illustrated in Figure \ref{fig:int_wiring}. Each of the three phases of the motors has to be connected to the motor outputs M0 and M1. The order in which the motor phases are connected is not important. The ODrive controller will figure it out during a calibration phase. However, after calibration, the order cannot be changed. If changed, consider re-calibrating the motors. 
      
      Furthermore, hoverboard motors are equipped with five hall sensors coloured red, black, blue, green and yellow for position feedback. Unfortunately, the ODrive controller has no noise-filtering capacitors and consequently, the hall sensors are susceptible to noise. To get consistent and clean readings from the hall-effect sensors, noise filtering capacitors are required to be connected to the corresponding ODrive's J4 pinouts as described in Table \ref{tab:8}.
    
    \begin{figure}
    \centering
    \includegraphics[scale=0.55]{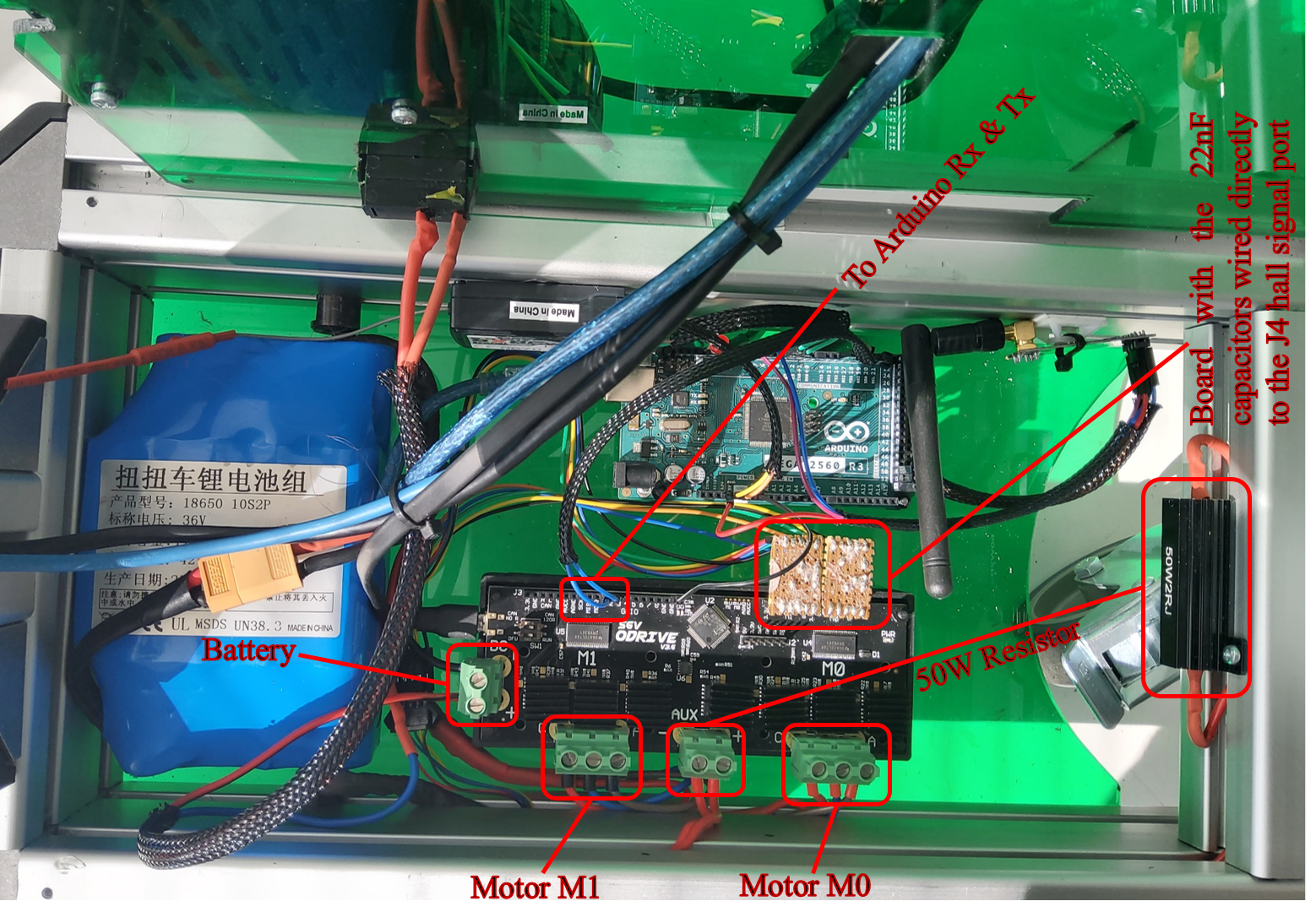}
    \caption{Internal wiring of the ODrive board. On the bottom, the three phases of the motors (M0 and M1) are connected. On the top, 22nF capacitors are used as noise filters for the hall sensors}
    \label{fig:int_wiring}
\end{figure}

   The required value of the filtering capacitor is approximately $22nF$. However, if you do not have exactly the required $22nF$, connecting two $47nF$ (c1 = c2 = 47nF) in series to obtain approximately 23.5nF will also work. Also, a 50W power resistor is required if the robot runs on a battery. This prevents the ODrive from unexpected shutdown as a result of regenerated energy into the battery. Usually, the resistors come along with the ODrive controller during supply. 
   
\arrayrulecolor{black}
\begin{table}[h]
\caption{Hall sensor wiring at the J4 signal port of the Odrive.}
\begin{center}
\begin{tabular}{p{0.10\linewidth} p{0.150\linewidth}}
  \toprule
    {\textbf{Hall wire}} & {\textbf{J4 signal port}}\\
     \midrule
    Red & 5V \\
    Yellow  & A \\
    Blue  & B \\
    Green  & Z \\
    Black  & GND \\
   \bottomrule
\end{tabular}
\end{center}
\label{tab:8}
\end{table}
 
 \item Arduino to ODrive interconnection: ODrive communicates with Arduino through a serial port or a universal asynchronous receiver/transmitter (UART). The UART pinouts are GPIO 1 (TX) and GPIO 2 (RX) which should be connected to the Arduino Arduino's RX (pin 18) and TX (pin 17) respectively. Also, there is a need to connect the grounds (denoted by GND) of the Arduino and the ODrive controller boards.

\item Power distribution: It is recommended to use two separate power sources for the ODrive controller and Jetson Nano to avoid a ground loop \cite{odrive2}. 36V lithium-ion 4400mAh battery is wired directly to the ODrive motor controller, and an additional power-bank battery ($P27$) was used to power the Jetson which requires only 5V 2500mA.

\item The nRF24L01+ module, MPU-9250 and Arduino connections: From Figure \ref{fig:gesture}, the control unit consists of $P17$, $P19$, and $P22$. These parts are required for the wireless transmission of data. The MPU-9250 and the Arduino are connected with four cables, the ground (GND), the power supply (VCC) and two cables for the I2C communication (SDA, SCL). The SDA pin of the MPU-9250 is connected to the SDA (pin 20) of the Arduino, and the SCL pin to the SCL (pin 21) of the Arduino. A power supply between 2.4V and 3.6V is needed. As a consequence, the 3.3V power supply pin of the Arduino is recommended to be used \cite{imu}. The communication between the nRF24L01+ module and the Arduino is established via an SPI interface. In Figure \ref{fig:nrf24l01}, a detailed description of the nRF24L01+ pinout can be seen. 

\begin{figure}
    \centering
    \includegraphics[scale=0.60]{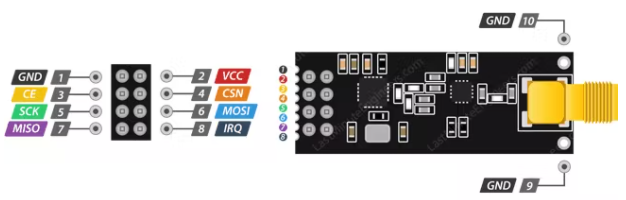}
    \caption{Illustrated is nRF24L01 module with its pinout \cite{7987982}. The module is used for the wireless communication between the Arduino at the remote control unit and the Arduino at the robot unit (see Figure \ref{fig:gesture}).}
    \label{fig:nrf24l01}
\end{figure}

The SPI pins of the Arduino (Mega 2560 Rev3) are MISO $\rightarrow$ pin 50; MOSI $\rightarrow$ pin 51, and SCK $\rightarrow$ pin 52. In this work, the pins are connected as follows: nRF2401 GND $\rightarrow$ Arduino GND; nRF2401 VCC $\rightarrow$ Arduino 3.3V; nRF2401 CE $\rightarrow$ Arduino digital 7; nRF2401 CSN $\rightarrow$ Arduino digital 8; nRF2401 MOSI $\rightarrow$ Arduino digital 51; nRF2401 SCK $\rightarrow$ Arduino pin 52; nRF2401 MISO $\rightarrow$ digital 50. Furthermore, at the robot unit (Figure \ref{fig:gesture}), the nRF24L01+ module and the Arduino are wired in the same way as the control unit.

\item RC receiver (P23) wiring to Arduino: RC receivers are needed to drive the motor using pulse-width modulated (PWM) signals \cite{dave}. A typical RC receiver such as the one used in this work (P23) has three kinds of pins, two of which are GND and 5V respectively, and the remaining are for PWM signals. Hoverboard motors require three signal pins, i.e., throttle, steering and enable \cite{lukas}. Therefore, the receiver to the Arduino wiring is described as follows:
RC\_GND $\rightarrow$ Arduino\_GND, RC\_5V $\rightarrow$ Arduino\_5V, RC\_CH1 $\rightarrow$ Arduino\_port2, RC\_CH2 $\rightarrow$ Arduino\_port3, and RC\_CH3 $\rightarrow$ Arduino\_port18.
\end{enumerate}

\section{Operation instructions}\label{subsec6}
To get started operating the $ROMR$ in real-time for the first time, the first step is to power ON the robot by pressing the ON-OFF switch beside the Jetson nano board ($P16$) and start controlling it with the transmitter. Note that after powering ON the robot, a red LED blinks. You have to wait until the blinking stops, then it is ready to be used. However, if you have changed the default configuration, or probably wish to rebuild and reconfigure the robot from scratch to operate it in other modes, then this section provides step-by-step instruction to get started. Before these steps,  make sure that all the robot parts are properly assembled and wired according to the instructions in Section \ref{sec5}.

\subsection{Initial configuration and setup instruction}
This section provides details about the initial configuration of the $ROMR$. It is recommended to follow the instruction in this section very careful as it determines how well the system would perform.
\subsubsection{Nvidia Jetson Nano set up and ROS installation}
 To set up the Nvidia Jetson Nano, some basic tools such as a microSD card (32GB minimum recommended), a USB keyboard and a mouse, a computer display (HDMI or DP) and a micro-USB power supply are required. The microSD card and micro-USB power supply usually come with the Jetson Nano during supply, if you purchased the full development kit. The setup instructions are as follows:
 
 \begin{enumerate}
     \item Download the Jetson Nano developer SD card image (JetPack), and write the image to the microSD card that usually come along with the Jetson Nano board. The reference instruction to configure the JetPack for the first time can be found at \url{https://developer.nvidia.com/embedded/learn/get-started-jetson-nano-devkit}.
     
     \item Install ROS and its packages on the JetPack. The instruction for the installation is provided at the official ROS wiki page at \href{http://wiki.ros.org/melodic/Installation/Ubuntu}{http://wiki.ros.org/melodic/Installation/Ubuntu} for ROS Melodic, which is supported per default by the JetPack. Or follow the instructions here: \url{https://github.com/Qengineering/Jetson-Nano-Ubuntu-20-image} to configure Ubuntu 20.04 OS image for ROS Noetic installation. Thereafter, install ROS Noetic from \url{http://wiki.ros.org/noetic/Installation/Ubuntu}.
     
    \item Create a workspace. A workspace is a set of directories with which you can store the ROS code that you may have written. Instructions can be found at \url{http://wiki.ros.org/catkin/Tutorials/create_a_workspace}.
 \end{enumerate}

\subsubsection{Setup the ODrive tool and calibrate the BLDC motors}\label{subsec53}
 To begin the initial configuration and calibration of the hoverboard BLDC motors, make sure that all the necessary wiring has been completed as described in subsection \ref{sec52}. Also, ensure that all relevant switches are turned ON, and the motors are positioned in such a way that they can freely move. The ODrive need to be connected to the host computer (the Nvidia Jetson Nano). ODrive has a python3 programming interface, called "odrivetool" for configuring the BLDC motors and commanding them to move at a specific number of revolutions per minute or rotations.
 
 Therefore, python3 is required to be installed first on the host computer before setting up the "odrivetool".  The instruction for the setup can be found at \href{https://docs.odriverobotics.com/v/latest/getting-started.html#install-odrivetool}{https://docs.odriverobotics.com}. Alternatively, you could simply download and run the calibration script which has been prepared to avoid the tedious task of following the tutorial to calibrate the motors at \href{https://osf.io/awf9t}{https://osf.io/awf9t}.
 The script will configure the axes of the motors and their respective encoders as well as set motor parameters such as the velocity gain, the position gain, the bandwidth, and more.
 After the successful calibration, you can test the motors from the "odrivetool" command line to ensure that it is properly configured and ready to receive velocity commands. First, start the "odrivetool" and from the command line send the following commands to the motors:
 
 \begin{lstlisting}[language = python]
 # Place the motor connected to axis0 (M0) of the ODrive board at velocity control mode.
     odrv0.axis0.controller.config.control_mode =  CONTROL_MODE_VELOCITY_CONTROL
 # Place the axis0 (M0) motor in closed-loop control mode.
     odrv0.axis0.requested_state = AXIS_STATE_CLOSED_LOOP_CONTROL 
 # At this point, the axis0 motor should spin at 1 turn/s.
     odrv0.axis0.controller.input_vel = 1
 # Stop the motor on axis0 from spinning (set the velocity to 0 turn/s).    
     odrv0.axis0.controller.input_vel = 0
 # Disable the motor on axis0 and return it to an idle state.    
     odrv0.axis0.requested_state = AXIS_STATE_IDLE
  \end{lstlisting}
 
 Repeat the same test with the second motor that is connected to axis 1 of the ODrive board. If the configuration and calibration of the motors are correct, then both motors should spin until they receive 0 as commanded velocities. Or receives idle state commands.
 At any point during the calibration and testing with the interactive "odrivetool", always use the code below to list calibration errors and to clear them.
 \begin{lstlisting}[language = python]
 dump_errorrs(odrv0,True) # dumps ODrive calibration errors and clear them
 \end{lstlisting}

\subsubsection{Install Arduino integrated development environment (IDE) and connect to ROS}
The Arduino IDE allows one to write software programs (sketches) and upload them to the Arduino board for robot control. The steps to set up the IDE, and connect it to ROS are described below:
\begin{enumerate}
     \item  First, download the Arduino IDE at \href{https://www.arduino.cc/en/Guide}{https://www.arduino.cc/en/Guide} and follow the onscreen instructions to set it up.
     \item Integrate the Arduino to communicate with the ROS via rosserial node. The rosserial\_arduino package enables the Arduino to communicate with the Jetson Nano via a USB-A male to USB-B male cable. The setup instruction can be found at the \href{http://wiki.ros.org/rosserial\_arduino/Tutorials/Arduino\%20IDE\%20Setup}{http://wiki.ros.org/rosserial\_arduino}. Note: It is advised to use a udev rule for the USB devices. This will allow the devices to be recognized and configured automatically when it is plugged in.
     \item  Launch the ROS serial server by running the code below at the command line to ensure that the setup was successful.
     \begin{lstlisting}[language = python]
     rosrun rosserial_python serial_node.py  /dev/ttyACM0  # rosserial python node
     \end{lstlisting}
    For a detailed explanation of the above rosserial-python node, visit the ROS wiki address at \url{http://wiki.ros.org/rosserial_python#serial_node.py}. Make sure that you check the port to which your Arduino is connected and the baud rate of your device. In our case, it is ttyACM0 and 115200 respectively. In your case, it may be different. Take note of it always.
     \item  Connect the Arduino to the ODrive. First, install the ODriveArduino library. Clone or download the repository \href{https://github.com/odriverobotics/ODrive/tree/master/Arduino}{https://github.com/odriverobotics/ODrive/tree/master/Arduino}. From the Arduino IDE, select Sketch $\rightarrow$ Include Library $\rightarrow$ Add .ZIP Library and select the enclosed zip folder.
     Run the \textit{"ODriveArduinoTest.ino"} sketch with the motors connected to ensure that it is properly configured and ready to accept commands. If everything went successfully, then the motors should move accordingly.
 \end{enumerate}

\subsubsection{The MPU-9250 and nRF24L01+ modules setup}
As shown in Figure \ref{fig:gesture}, the Arduino at the control unit reads the MPU-9250 data and forwards it to the robot via the nRF24L01+ module. To run the Arduino sketch, the "FaBo 202 9Axis MPU9250" library by Akira Sasaki released under the Apache license, version 2.0 must be installed, as well as the "rf24" library by TMRh20 Avamander released under the GNU general public license. The FaBo 202 9Axis MPU9250 library is used for reading the data measured by the MPU-9250 sensor, while the rf24 library is needed for the usage of the nRF24L01+ module. Both libraries can be found in the library manager of the Arduino IDE.


The Arduino at the robot unit primarily forwards the raw MPU-9250 data that it receives from the nRF24L01+ module to the Jetson Nano. This is done by publishing the data to the "imu/data\_raw" topic of the ROS system, which is running on the Jetson Nano. The Jetson Nano converts the raw data into velocity commands that include a target linear velocity and a target rotation speed of the robot.
The Arduino receives these velocity commands by subscribing to the "cmd\_vel" topic. After the Arduino receives the velocity commands, it converts them into target speed values of the motors that are set on the ODrive board.

 \subsubsection{Setup the RPlidar}\label{sec65}
 The RPlidar sensor provides the scan data required for mapping, localization and navigation purposes. It is connected to the Jetson Nano or the host computer through a USB serial port. The procedure for its setup is summarized in the following steps.
 \begin{enumerate}
     \item Clone the RPlidar ROS packages \href{https://github.com/Slamtec/rplidar_ros}{https://github.com/Slamtec/rplidar\_ros} to your ROS catkin workspace source directory and run the code below to build the rplidarNode and rplidarNodeClient.
    \begin{lstlisting}[language = XML]
    catkin_make 
   \end{lstlisting} 
     \item Check the authority of the rplidar serial port by typing at the command window:
     \begin{lstlisting}[language = XML]
    ls -l /dev |grep ttyUSB
   \end{lstlisting}
     Take note of the port in which the USB is connected e.g., .../ttyUSB0. Add authority to write the USB:
    \begin{lstlisting}[language = XML]
    sudo chmod 666 /dev/ttyUSB0
   \end{lstlisting}
     \item Launch the RPlidar node to view and test if the setup was successful.
    \begin{lstlisting}[language = XML]
    roslaunch rplidar_ros view_rplidar.launch
   \end{lstlisting} 
 If correctly set up, you will obtain an output similar to the one displayed in Figure \ref{fig:lab_gazebo}b with the lidar scan represented as red dots.
 \end{enumerate}
After the above setups, the $ROMR$ is ready for experimentation.

\subsection{Experimentation \& remote operation instruction}\label{sec60}
To allow a human operator to intuitively control the $ROMR$, we developed three remote control techniques. In this section, we provide step-by-step instructions on how these techniques can be implemented.

\subsubsection{ROMR teleoperation from remote-control (RC) devices}\label{sec731}
As per default, $ROMR$ is configured to operate in RC mode. However, if the default configuration has been altered or changed, then, the following steps must be taken to reconfigure it. Before these steps, make ensure that the ODrive controller has already been calibrated and configured to accept commands (see sub-subsection \ref{subsec53} for instructions). Also, it is important to ensure that the RC receiver is properly wired according to the instructions in subsection \ref{sec52}.
\begin{enumerate}
    \item Upload the \textit{"romr\_remote\_control.ino"} sketch to Arduino. Before that, the "Metro" library has to be included in the Arduino IDE library. The "Metro" library can be downloaded from \url{https://github.com/thomasfredericks/Metro-Arduino-Wiring}.
    \item With the motors switched off, move the RC transmitter sticks and monitor it from Arduino serial plotter. If there is communication between the receiver and the transmitter, you would obtain a similar response as the one shown in Figure \ref{fig:rc_test} from the Arduino serial plotter.
    \begin{figure}
    \centering
    \includegraphics[width=0.55\textwidth]{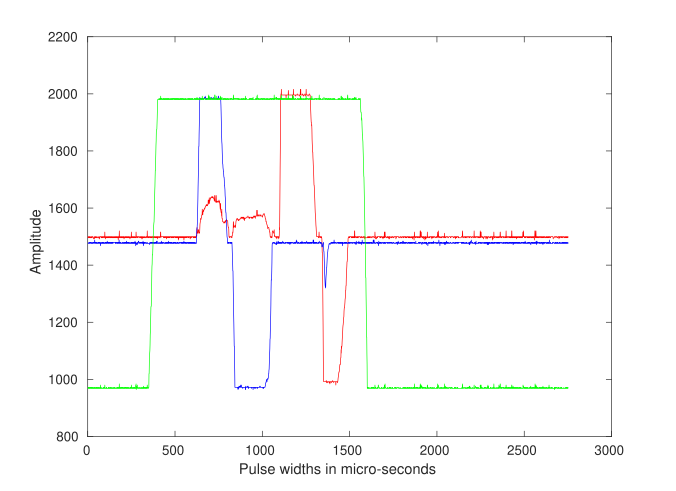}
    \caption{ Setting up the RC control on the Turnigy 9x. The red line indicates PWM activation, the blue line indicates throttle control and the green line indicates steering control.}
    \label{fig:rc_test}
 \end{figure}
 \end{enumerate}

\subsubsection{ROMR control from Android-based device}\label{sec732}
\begin{figure}[h]
    \centering
    \subfigure[]{\includegraphics[width=0.55\textwidth]{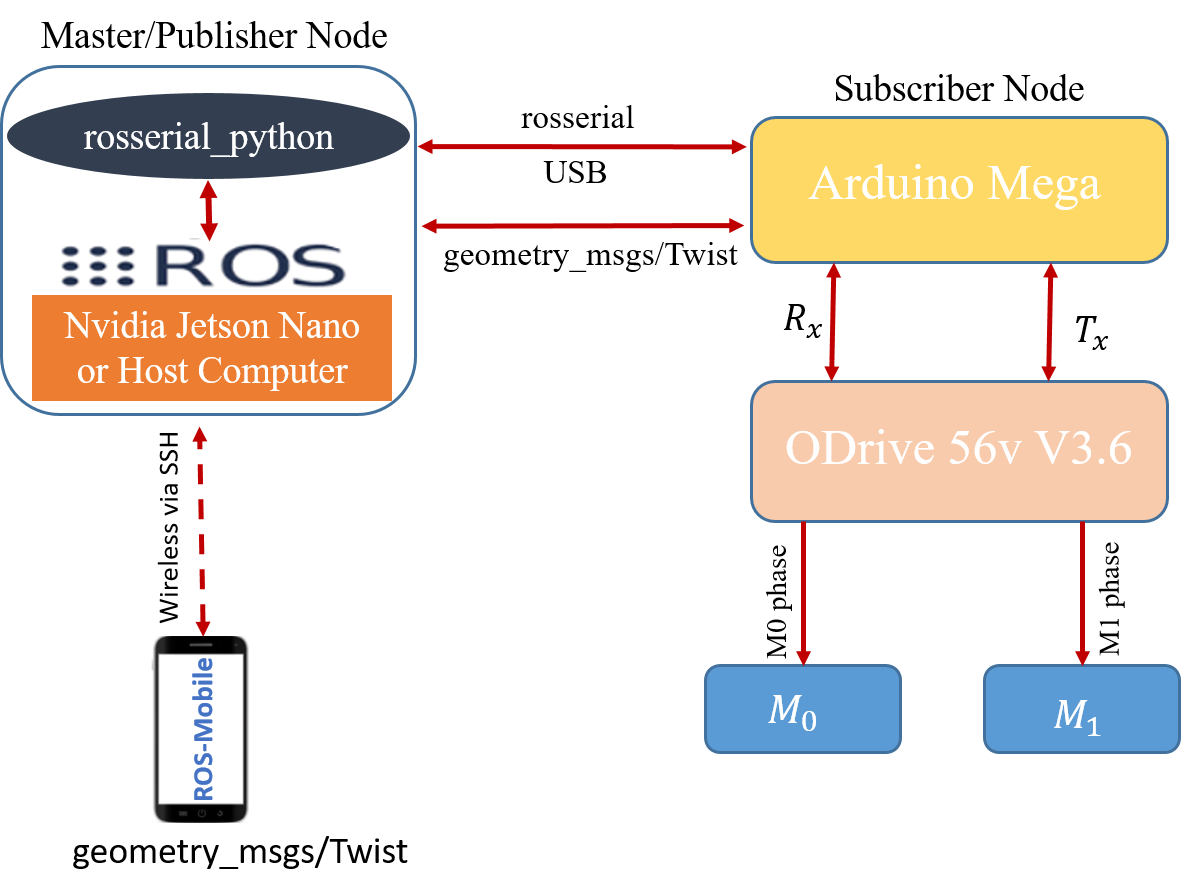}} 
    \subfigure[]{\includegraphics[width=0.4\textwidth]{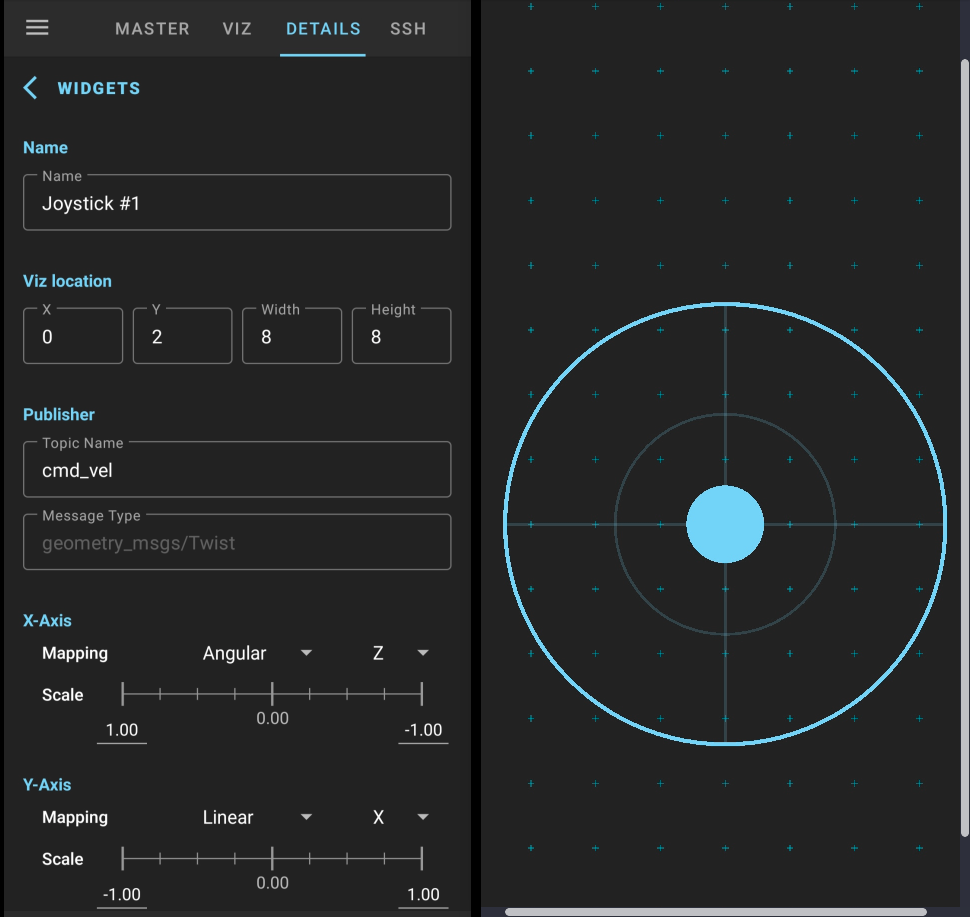}} 
    \caption{ The $ROMR$ real-time control and monitoring with ROS-Mobile device (a) System communication structure (b) Control and monitoring from ROS-Mobile \cite{rottmann2020ros} or Android-based devices.}
    \label{fig:ros-mobile}
\end{figure}

One of the main features of the $ROMR$ is the ability to be teleoperated from any Android-based device. The idea is to alleviate the need for complex robot teleoperation devices such as a joystick, an RC transmitter, etc., and to provide an intuitive way of controlling the robot by simply touching the Android device screen. To achieve this, we leveraged the framework developed by Rottmann Nils et al. \cite{rottmann2020ros}. The setup is straightforward. First, you have to make sure that Arduino has been set up to communicate with ROS. If you have not done that yet, it is advisable to follow the instructions in Section \ref{subsec6}. "rosserial" is very important for this section. Therefore, make sure that the "rosserial" python node (rosserial\_python serial\_node.py) is running properly. The whole communication structure is described in Figure \ref{fig:ros-mobile}a.

As illustrated in Figure \ref{fig:ros-mobile}, the "rosserial" python node allows all the compatible connected electronics to communicate directly with the robot using the ROS topics and messages. All the information between the interconnected systems is communicated with the help of the rosserial package. Make sure that the robot is switched ON, the battery is connected, and the wheels are free to spin. If you have changed the default ODrive calibration, make sure that the calibration is completed before continuing. Open the downloaded ROS-Mobile App, which enables ROS to control the robot's joint velocities. The App supports linear (forward and backward movement) and angular (rotation around the z-axis) movements. See Figure \ref{fig:ros-mobile}b for the setup. An SSH connection has to be  established between the devices, and all the devices have to be on the same wireless network. The steps are summarised as follows:

\begin{enumerate}
    \item Download the ROS-Mobile App from the Google Playstore.
    \item Configure the IP address. First, connect the robot and the Android device to the same wireless network. From the command window terminal, type "ifconfig", this will display the IP address, e.g., 192. 168.1.15.
    \item At the "master" node URI of the ROS-Mobile App, enter the IP address and 11311 for the "master" port. Ensure that \textit{roscore} is running, and click on the connect button.
    \item Once the above steps are completed, upload the \textit{"ros\_mobile\_control.ino"} sketch to Arduino. While the \textit{roscore} is still running, run the rosserial python node in a separate terminal: 
    \begin{lstlisting}
    rosrun rosserial_python serial_node.py /dev/ttyACM0
    \end{lstlisting}
    Take note of the .../dev/ttyACM0 Arduino port. It may be different in your case.
    \item At the "details" tab of the ROS-Mobile App, select "Add widget", select "joystick" and set the XYZ-coordinates accordingly. Click on the "viz" tab to visualise and control the robot. Ensure that the rosserial python node is running and the cmd\_vel topic is been subscribed to. Once done, the robot can be controlled by simply touching the respective coordinates on the screen.
\end{enumerate}

\subsubsection{Gesture-based control of the ROMR}\label{sec733}
Unlike the traditional or "ready-to-use" robot control approaches such as a joystick or the ROS rqt plugin, a gesture-based approach has the potential to control the robot in a very intuitive way. This strategy allows the operator to focus on the robot instead of the controller. The goal is for a non-robotics expert to be able to remotely navigate the robot depending on the direction in which the operator’s hand is tilted. For example, if the hand is tilted forward (pitch angle), the robot should move forward. If the hand is tilted to the side (roll angle), the robot should rotate. Since the MPU-9250 sensor does not measure the orientation of the operator’s hand, but only the acceleration and the rotations speed of the sensor, the IMU sensor ($P19$) data must be fused to determine the orientation of the operator’s hand and the tilt angle of the sensor.

\begin{figure}
    \centering
    \subfigure[]{\includegraphics[height=0.24\textheight, width=0.22\textwidth]{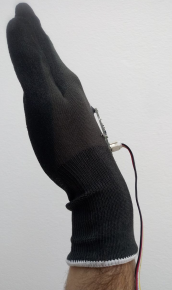}}
    \subfigure[]{\includegraphics[height=0.24\textheight, width=0.22\textwidth]{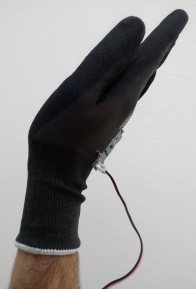}}
    \subfigure[]{\includegraphics[height=0.24\textheight, width=0.24\textwidth]{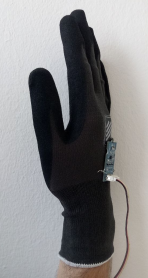}}
    \subfigure[]{\includegraphics[height=0.24\textheight, width=0.24\textwidth]{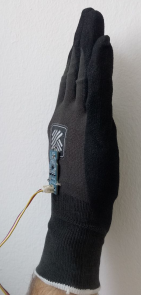}}
    \caption{ $ROMR$ real-time control and monitoring based on hand movement (a) hand tilt forward $\rightarrow$ the robot moves forward (b) hand tilt backwards $\rightarrow$ the robot moves back (c) hand tilt to the right $\rightarrow$ the robot rotates in a clockwise direction (d) hand tilt to the left $\rightarrow$ the robot rotates in a counter-clockwise direction.}
    \label{fig:hand_gesture}
\end{figure}

We implemented four gestures with different hand motions (see Figure \ref{fig:hand_gesture}). We used the IMU sensor ($P19$) to measure the hand gestures needed and map them onto the robot's linear and angular velocities. The data collected by the IMU sensor is sent via a wireless connection to the robot, which uses it to calculate the movement commands for the motors. We leveraged the framework proposed in \cite{Guanhao} and \cite{unknown} to achieve the gesture-based control strategy. 

\begin{figure}[h]
    \centering
    \includegraphics[scale=0.50 ]{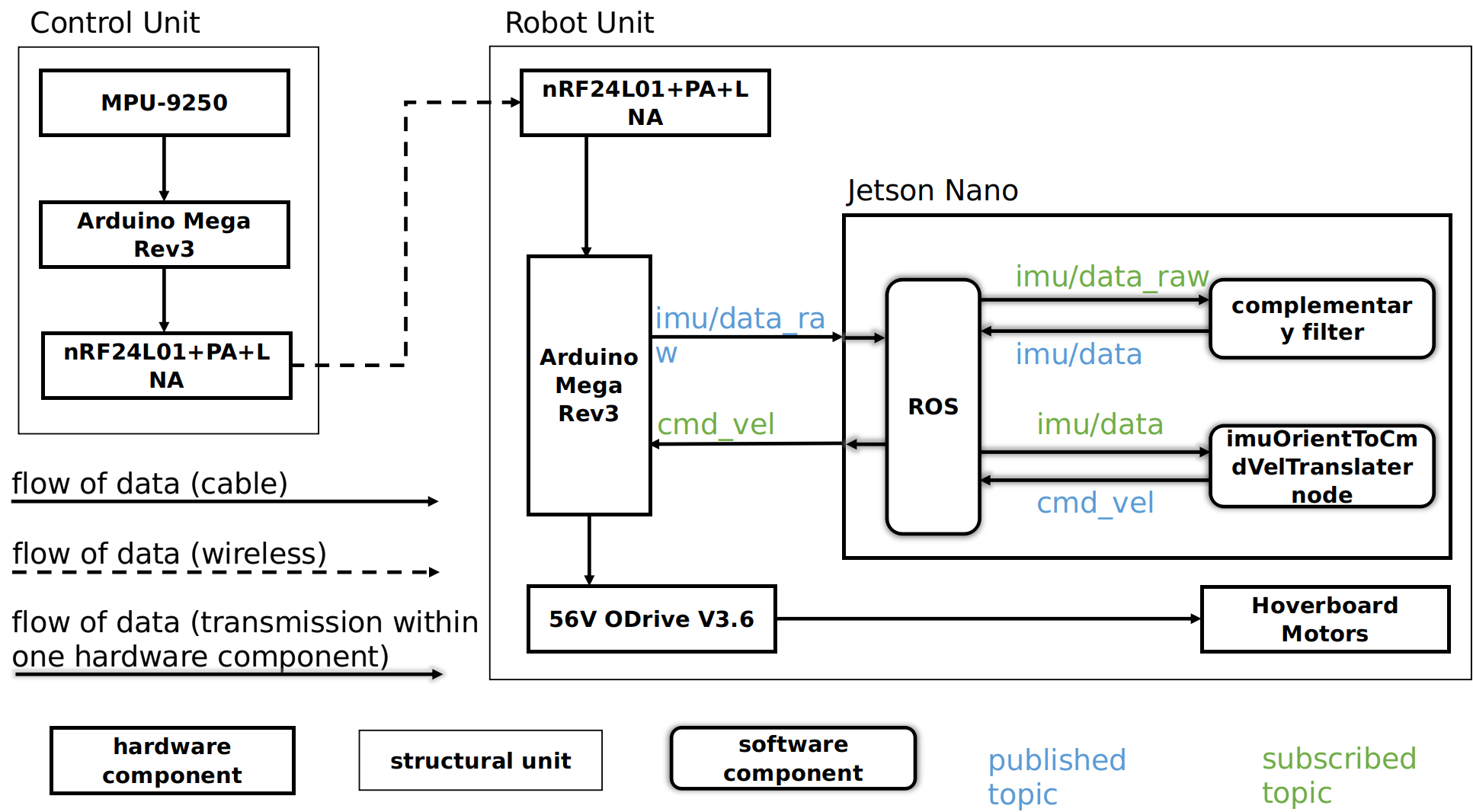}
    \caption{ Block diagram illustrating the architecture and flow of information during the gesture control approach.}
    \label{fig:gesture}
\end{figure}

A detailed overview of the hardware and software architecture and the data flow is shown in Figure \ref{fig:gesture}. The components are divided into a control unit and a robot unit, as the components of these two units are physically located in different places. While the control unit is been attached to the user's arm/hand, the robot unit is within the $ROMR$ board.

The step-by-step instruction to operate the robot based on gesture demonstration is described as follows:
\begin{enumerate}
    \item At the control unit, upload the \textit{"IMUDataNRF24L01\_Transmitter.ino"} sketch to Arduino, and at the robot unit, upload \textit{"NRF24L01Receiver\_PC\_WheelController\_WheelMonitoring.ino"} sketch to the Arduino.
    \item From the host computer, run \textit{roscore} to start the ROS master, and then: 
    \begin{lstlisting}[language = python]
    roslaunch romr_robot romr_bringup.launch # Or
    rosrun rosserial_python serial_node.py /dev/ttyACM0
    \end{lstlisting} to start rosserial python node and establish a connection between the Arduino and the Jetson Nano. Depending on which port the Arduino is connected to the Jetson Nano, the port ttyACM0 must be adjusted accordingly.
    \item Finally, execute the following script to start the complementary filter node:
    \begin{lstlisting}[language = C]
    rosrun romr_robot imuOrientToCmdVelTranslater
    \end{lstlisting}
    By switching the robot ON, the motors are automatically powered and ready to be controlled with the IMU sensor. Depending on the direction the IMU sensor is tilted, a corresponding movement of the robot is obtained as shown in Figure \ref{fig:hand_gesture}a-d. During the operation, attention must be paid so that the hand is not tilted about $90^\circ$.
\end{enumerate}

\section{Validation and characterization}
$ROMR$ has been successfully developed, tested and validated both in simulation and in real-world scenarios. Experiments were performed to characterise its performance, robustness and suitability for research, navigation and logistics applications. The evaluation results are presented in this section. Furthermore, the validation video can be viewed at \url{https://osf.io/ku8ag}.

\subsection{Simulation scenarios}\label{subsec:41}
Although the development of $ROMR$ focused on real-world applications, it is also important to have a 3D simulation model of the robot, to enable users to work in virtual environments to explore tools, techniques and methods. Furthermore, the simulation model could also enable the $ROMR$ users to become familiar with 3D simulation and visualisation tools such as Gazebo \cite{gazebo} and Rviz \cite{rviz}. 

\begin{figure}
    \centering
    \subfigure[]{\includegraphics[height=0.23\textheight, width=0.40\textwidth]{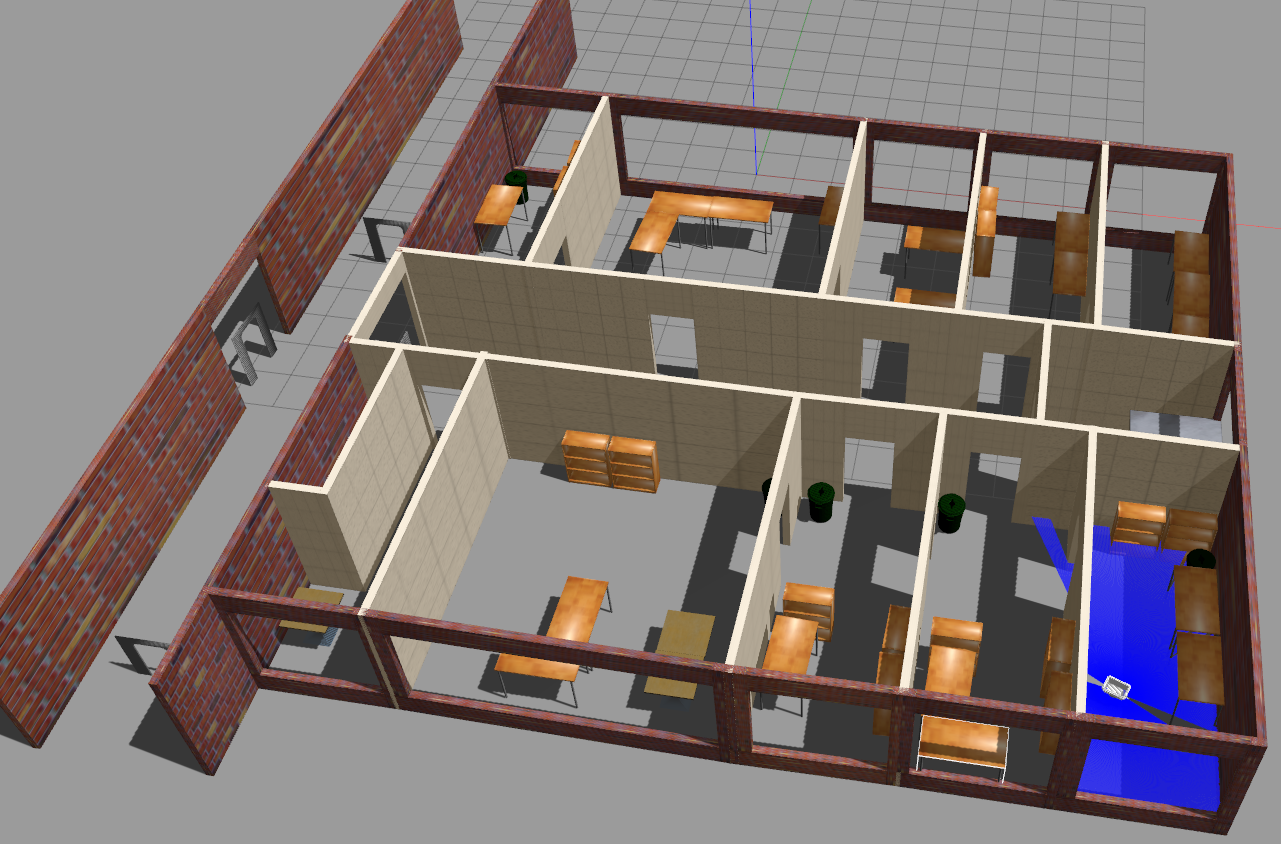}} 
    \subfigure[]{\includegraphics[height=0.23\textheight, width=0.55\textwidth]{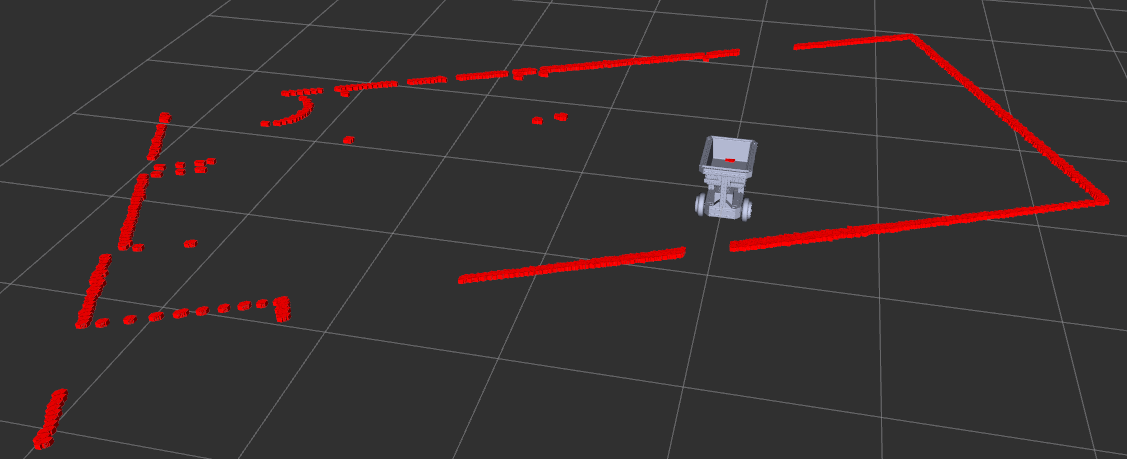}} 
    \caption{ $ROMR$ simulation platform.  (a) The operational environment model with the robot models together with the sensor model. The blue lines are the lidar sensor scan of the operational environment in the Gazebo. (b) Rviz visualisation. The red dotted lines are the lidar scan showing the location of obstacles (or objects) within the robot environment.}
    \label{fig:lab_gazebo}
\end{figure}

The simulation platform includes three parts: the environment model, the robot model, and the sensors model. For the environmental model, we created the floor plan of our laboratory environment using the Gazebo model editor (see Figure \ref{fig:lab_gazebo}a). 
Taking advantage of the ROS framework\cite{ros}, the $ROMR$ model was implemented in accordance with the unified robot description format (URDF) \cite{urdf}. The URDF is an extended mark-up language (XML) format that describes all kinematic and dynamic properties of the robot, the physical elements, such as the links, the joints, the actuators, and the sensors\cite{8324585}.

We generated the URDF of the robot including the sensors model using the 3D CAD models described in Table \ref{tab:5}. Further, all the models were verified with several simulation tests as depicted in Figure \ref{fig:lab_gazebo}. The step-by-step procedure for this simulation is as follows:
\begin{enumerate}
    \item Download the $ROMR$ ROS files at \url{https://osf.io/e4syc} to your catkin workspace and build it.
    \item Open three terminal windows, and execute the following in each of the terminals:  
    \begin{lstlisting}[language = C]
    roscore 
    \end{lstlisting}
    \begin{lstlisting}[language = C]
    roslaunch romr_robot romr_house.launch
    \end{lstlisting}
    to launch the $ROMR$ world (operational environment model in Gazebo \cite{gazebo}) with the robot spawned and the sensor model active (see Figure \ref{fig:lab_gazebo}a). At the third terminal run the following node to visualise in Rviz (see Figure \ref{fig:lab_gazebo}b).
     \begin{lstlisting}[language = C]
    roslaunch romr_robot romr_rviz.launch
    \end{lstlisting}
    \item Navigate the robot within the operational environment using the ROS rqt plugin, keyboard or the framework described in sub-subsection \ref{sec732}.
\end{enumerate}

\subsection{General system validation test}
This section provides information about the robustness of the system by completing different tests with the robot and its sub-components. The outcome of each test is presented in Table \ref{tab:load_test}.

\arrayrulecolor{black}
\begin{table}[ht]
\caption{General systems validation test.}
\begin{center}
\begin{tabular}{p{0.15\linewidth} p{0.20\linewidth} p{0.20\linewidth} p{0.20\linewidth} p{0.11\linewidth}}
  \toprule
    {\textbf{Test name}} & {\textbf{Test purpose}} & {\textbf{Test Process}} & {\textbf{Expected Result}} & {\textbf{Outcome}}\\
\midrule
    Chassis test & To Verify the solidity and stability of the robot's chassis & Complete five different movement tasks at high speed with all the components mounted & All the subsystems must be stable and rigid throughout the test & Passed \\
\midrule
   Payload test & To Verify the maximum load capacity the robot can carry without affecting the controller & Place different loads on the robot and check the response & The robot should be able to convey the load up to the maximum capacity & Carried up to 90kg (see Figure \ref{fig:load_test}) \\
\midrule
   Reconfiguration test & To Verify how long and easy to dismantle and re-assemble the system  & Dismantle all the subsystems including the chassis and wiring and re-assemble them. Record the time taken to complete the process & Should not take more than 3 hours & Took about 1 hour 15 minutes \\
\midrule
    Battery live test  & To evaluate how long the battery can power the robot for a long mission task & Operate the robot continuously with all the electronics parts active for a long period of time  & The battery should last up to the maximum capacity & The battery lasted for about 8 hours \\
\midrule
    $ROMR$ stability test  & To verify the stability of the robot when driven at high speed and at small turning radii & Drive the robot forward and backwards, at high speed, and in a circular path of different radii. Record the linear and angular velocity data & The robot should be stable throughout the stability test process  & See Subsection \ref{stab} and Figure \ref{fig:stab_test} for the results \\
 \bottomrule
\end{tabular}
\end{center}
\label{tab:load_test}
\end{table}

\subsubsection{Payload capacity test}

\begin{figure}[h]
    \centering
    \subfigure[]{\includegraphics[height=0.21\textheight, width=0.19\textwidth]{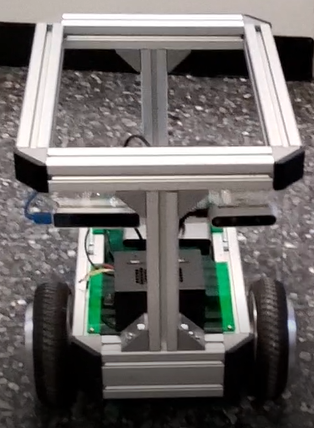}} 
    \subfigure[]{\includegraphics[height=0.21\textheight, width=0.21\textwidth]{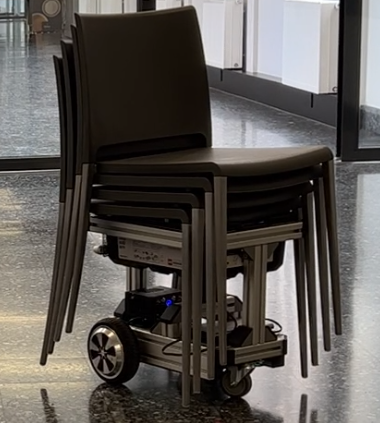}}
    \subfigure[]{\includegraphics[height=0.21\textheight, width=0.20\textwidth]{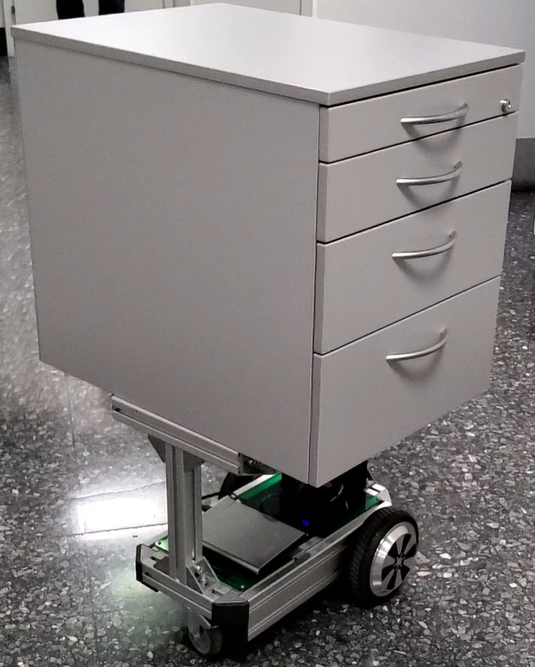}} 
    \subfigure[]{\includegraphics[height=0.21\textheight, width=0.17\textwidth]{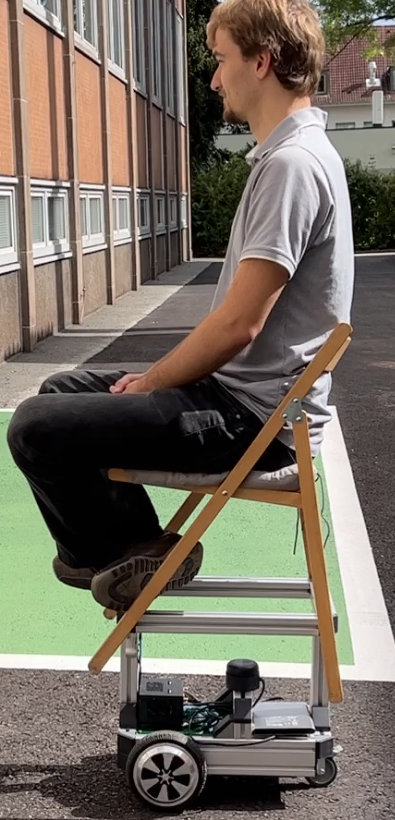}}
    \subfigure[]{\includegraphics[height=0.21\textheight, width=0.17\textwidth]{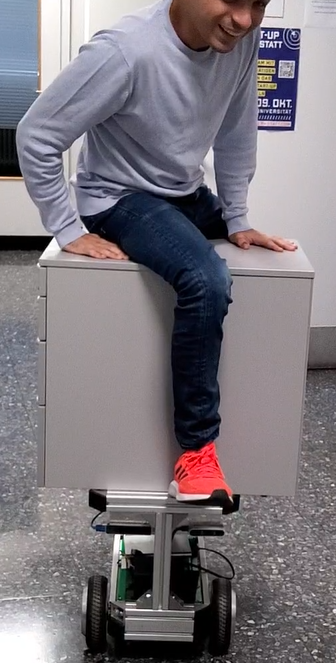}}
    \caption{ Validation of the maximum load capacity of the $ROMR$. Shown are (a) $ROMR$ weight only ($17.1\;kg$), (b)  $16\;kg$, (c) $25\;kg$, (d) $85\;kg$, and (e) $90\;kg$.}
    \label{fig:load_test}
\end{figure}

In Figure \ref{fig:load_test}, the result of the payload capacity test carried out as described in Table \ref{tab:load_test} is presented. The robot is teleoperated to move along linear and angular trajectories with different loads ranging from the robot weight only (17.1kg) to 90kg. The goal is to verify how much load the robot can carry without affecting the controller.
Although the $ROMR$ can carry a load up to a maximum of 90kg, it is, however, not recommended to operate it at maximum load continuously to extend its life span. All the various load tests were carried out on the robot while moving on a flat surface. Thus, we did not evaluate the performance on irregular or unstructured surfaces.

\subsubsection{Stability test}\label{stab}
We performed this test to evaluate the stability of the robot under high speed and different turning radii. We tested the robot stability in both Gazebo simulation and in real-world at four different speeds: $0.5\;m/s$, $1.0\;m/s$, $1.5\;m/s$, and $ 2.5\;m/s $. For each speed, we tested the robot at different turning radii ($0.5\;m, 1.0\;m, 1.5\;m, 2.0\;m,$ and $2.5\;m$), three different payloads ($ROMR$ weight, $25\;kg,$ and $85\;kg$), and three different positions of the centre of gravity ($- 0.1\;m, 0.0\;m,$ and $0.1\;m$). Note, the positions of the centre of gravity (pCOG) are defined relative to the midpoint on the ROMR's base frame. The centre of gravity (COG) is slightly altered by adjusting the loads (25kg and 85kg) at positions $0.1\;m$ (front), $-0.1\;m$ (behind), and $0.0\;m$ (COG unaltered) from the midpoint of the ROMR base frame. The results are summarized in Table \ref{tab:effect_of_turning_radius}.

\begin{figure}[h]
    \centering
    \subfigure[]{\includegraphics[width=0.325\textwidth]{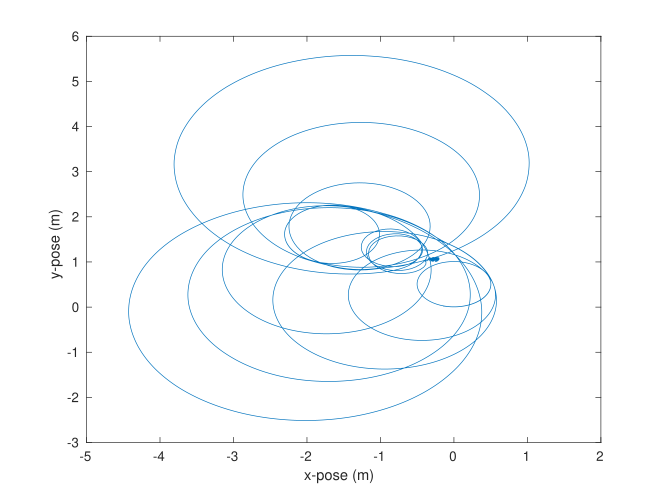}} 
    \subfigure[]{\includegraphics[width=0.325\textwidth]{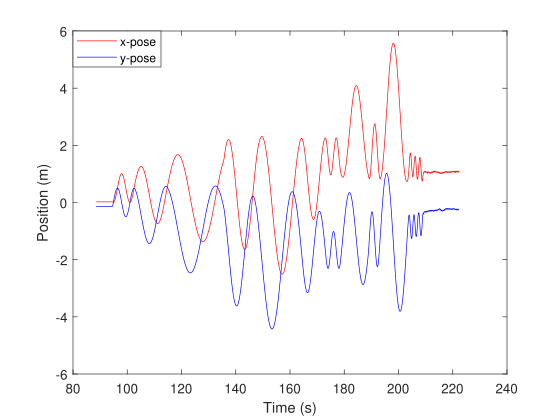}}
    \subfigure[]{\includegraphics[width=0.325\textwidth]{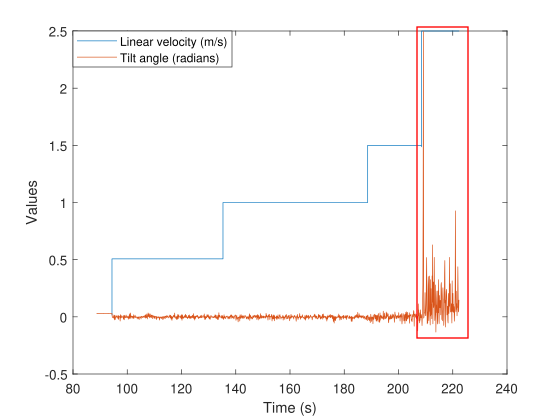}} 
    \subfigure[]{\includegraphics[width=0.325\textwidth]{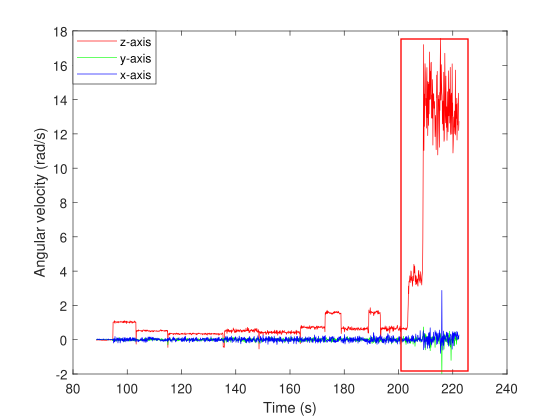}}
    \subfigure[]{\includegraphics[width=0.325\textwidth]{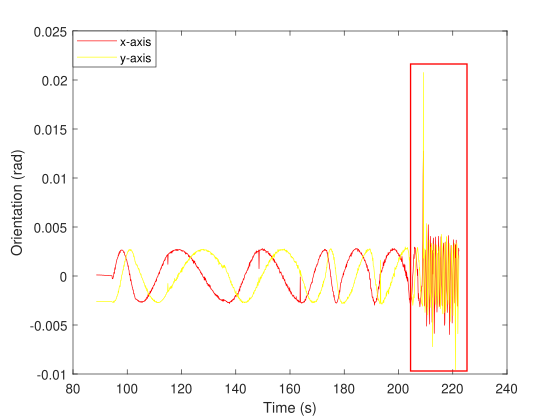}}
    \subfigure[]{\includegraphics[width=0.325\textwidth]{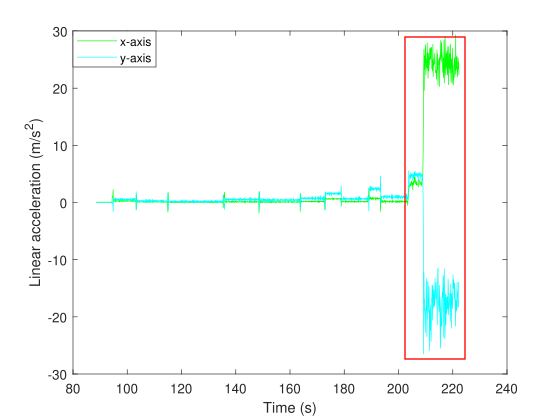}}
    \caption{The $ROMR$ stability test. The robot was controlled to follow circular paths at different linear velocities, turning radii, payload weights, and positions of the centre of gravity.  As shown in the sub-figures c - f, the robot was unstable at about $215$ seconds when the linear velocity increased to $2.5\; m/s$ (see the red rectangles). (a) depicts the robot's trajectory as it follows the circular path. (b) shows the robot's position on the  $x$ (red) and  $y$ (blue) axes at each time stamp. (c) shows the linear velocity in $m/s$ (blue) and the roll angle in $radians$ (dark red) at each time stamp. (d) shows the angular velocity of the robot along the $x$ (blue), $y$ (green) and $z$ (red) axes at each time stamp. (e) is the orientation at $x$ (red) and $y$ (yellow) axes respectively. (f) is the linear acceleration in $x$ (green) and $y$ (cyan) axes.}
    \label{fig:stab_test}
\end{figure}

\begin{table}[h]
 \caption{Minimum turning radius of the robot for different linear velocities, payload weights, and positions of the centre of gravity (pCOG).}
 \centering
\begin{tabular}{c|c|c|c|c}
\hline
 {\textbf{Linear Vel. (m/s)}} & {\textbf{Payload (kg)}} & {\textbf{pCOG (m)}} & {\textbf{Turning radius (m)}} & {\textbf{Stability}}\\
\hline
 0.5 & $ROMR$ weight (17.1) & 0.0 & 0.5 & stable\\
 0.5 & 25 & 0.1 & 1.0 & stable\\
 0.5 & 85 & - 0.1 & 1.5 & stable\\
\hline
1.0 & $ROMR$ weight (17.1) & 0.0 & 2.0 & stable\\
1.0 & 25 & 0.1 & 2.5 & stable\\
1.0 & 85 & - 0.1 & 1.5 & stable\\
\hline
1.5 & 25 &  - 0.1 & 1.0 & stable\\
1.5 & 85 & 0.1 & 2.5 & stable\\
1.5 & $ROMR$ weight (17.1) & 0.0 & 1.5 & stable\\
\hline
2.5 & 25 &  - 0.1 & 1.5 & stable\\
2.5 & 85 &  0.1 & 2.5 & stable\\
2.5 & $ROMR$ weight (17.1) & 0.0 & 0.5 & unstable\\
\hline
\end{tabular}
\label{tab:effect_of_turning_radius}
\end{table}
Figure \ref{fig:stab_test} shows the results of the stability test. The robot was driven in a circular path at different linear velocities, turning radii, payload weights, and pCOG. 
The odometry, command velocity and IMU data were recorded in a rosbag file used for our analysis. At the lowest speed of $0.5\;m/s$, the robot showed some minor oscillations when turning at the smallest turning radius of $0.5\; m$. However, these oscillations were not significant enough to cause the robot to lose control or become unstable. At the higher velocities of $1.0\; m/s$ and $1.5\;m/s$, the robot remained stable even when turning at the smallest radius of $0.5\;m$.  The robot was stable up to a linear velocity of $2.5\;m/s$. At $2.5\;m/s$, the robot started to show signs of instability, such as tilting and sliding. Therefore, it can be concluded that the maximum stable linear velocity of the robot is $2.5\;m/s$. 

To conduct the stability test, the following steps should be taken:
 
 \begin{enumerate}
     \item Launch the robot in the Gazebo environment:
     \begin{lstlisting}[language = python]
    roslaunch romr_robot  gazebo.launch # launches the robot in Gazebo.
    \end{lstlisting}
    \item The following nodes are needed only when launching the robot in the real-world. For the Gazebo simulation, they are not necessary.
    \begin{lstlisting}[language = python]
    roslaunch romr_robot  romr_bringup.launch # starts the rosserial python node.
    \end{lstlisting}
    \begin{lstlisting}[language = python]
    rosrun romr_robot  odomtorobottfgenerator # odometry broadcaster.
    \end{lstlisting}
    \item Start the stability test by launching the following node. The robot trajectory and a CSV file containing the necessary data for further analysis will be generated at the end of the test.
    \begin{lstlisting}[language = python]
    rosrun romr_robot  stability_test.py  # drives the robot at the specified velocities and turning radii.
    \end{lstlisting}
    \item If needed, record the IMU, the odometry, and the velocity data in a rosbag file for further analysis:
    \begin{lstlisting}[language = python]
    rosbag record -O stability_test.bag /imu /tf /odom /cmd_vel  # records relevant data for further analysis.
    \end{lstlisting}
 \end{enumerate}
 A video showing the result of the above test can be found at \url{https://osf.io/wcd4n}.
 
\subsection{Application of the ROMR for SLAM}\label{slam}
As stated earlier, $ROMR$ provides a framework for evaluating and developing SLAM algorithms. The SLAM problem is usually to build a map of an unknown environment, i.e, mapping while simultaneously keeping track of the estimates of a robot's pose (the position x, y and the orientation). Given a series of control and sensor observations \cite{slam}, \cite{Rueckert2010}, the map is built and the pose is estimated. We evaluated the Hector-SLAM algorithm \cite{hector-slam} on our RPlidar sensor ($P25$) to build the map of our real laboratory environment. The adaptive Monte Carlo localization (AMCL) \cite{amcl} approach was used to localise the robot within the built map. The advantage of the Hector-SLAM technique over other 2D SLAM techniques such as Gmapping \cite{gmapping}, and Google Cartographer \cite{cartmap} is that it only requires laser scan data and does not need odometry data to build the map. To generate the map, the following steps have to be followed:

\begin{enumerate}
    \item Set up the RPlidar as described in sub-subsection \ref{sec65}.
    \item Download or clone the Hector-SLAM packages at \url{https://github.com/tu-darmstadt-ros-pkg/hector\_slam.git} to your ROS workspace, and set the coordinate frame parameters according to the instruction at the ROS wiki page  \href{http://wiki.ros.org/hector_slam/Tutorials/SettingUpForYourRobot}{http://wiki.ros.org/hector\_slam}. Build the ROS workspace including the Hector-SLAM and RPlidar packages (\textit{catkin\_make}), then proceed to the next step.
    \item Open four terminal windows, and run the following in each of the terminal windows:
    \begin{lstlisting}[language = python]
    roscore # starts the roscore node
    \end{lstlisting}
    \begin{lstlisting}[language = python]
    roslaunch rplidar_ros rplidar.launch # launches the rplidar node
    \end{lstlisting}
    \begin{lstlisting}[language = python]
    roslaunch hector_slam_launch tutorial.launch # launches the hector-slam algorithm
    \end{lstlisting}
    While all the nodes are running, navigate the robot around the environment by employing any of the control approaches implemented in sub-subsections \ref{sec731}, \ref{sec732}, and \ref{sec733}. While building the map, it is recommended to move at a low speed such that a quality map is created.
    \item After the mapping is completed, execute the following at the fourth terminal:
    \begin{lstlisting}[language = python]
    rosrun map_server map_saver -f laboratory_map # saves the built map
    \end{lstlisting}
  Take note of the location where the map is saved. It would be required for localisation and autonomous navigation. The saved map can be viewed on your screen by running:
  \begin{lstlisting}[language = python]
    rosrun map_server map_server laboratory_map.yaml # view the saved map
    \end{lstlisting}
\end{enumerate}

\begin{figure}
    \centering
    \subfigure[]{\includegraphics[width=0.40\textwidth]{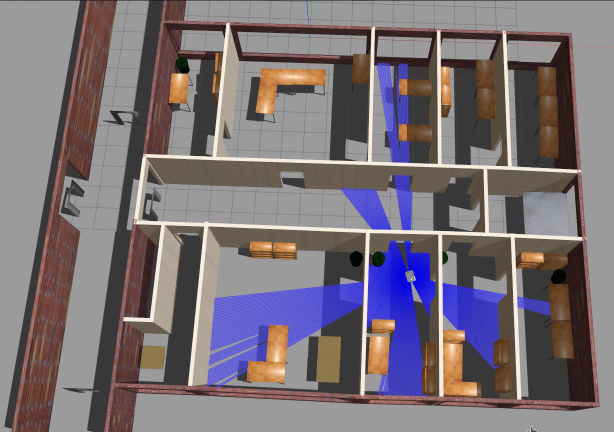}} 
    \subfigure[]{\includegraphics[width=0.54\textwidth]{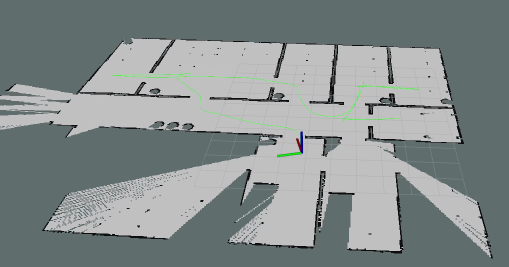}}
    \caption{ Generation of the 2D occupancy grid map of the environment using the 360$^\circ$ lidar sensor ($P25$) and a "Hector-SLAM" algorithm \cite{hector-slam}. (a) Virtual laboratory world in Gazebo. The blue lines represent the lidar scan. (b) Occupancy grid map of the operational environment in Rviz. The pale grey areas indicate the unoccupied (free) spaces that the robot can navigate, the black lines represent occupied areas not transversal by the robot, and the green line represents the robot's trajectory.}
    \label{fig:mapping}
\end{figure}

\begin{figure}[h]
    \centering
    \subfigure[]{\includegraphics[width=0.48\textwidth]{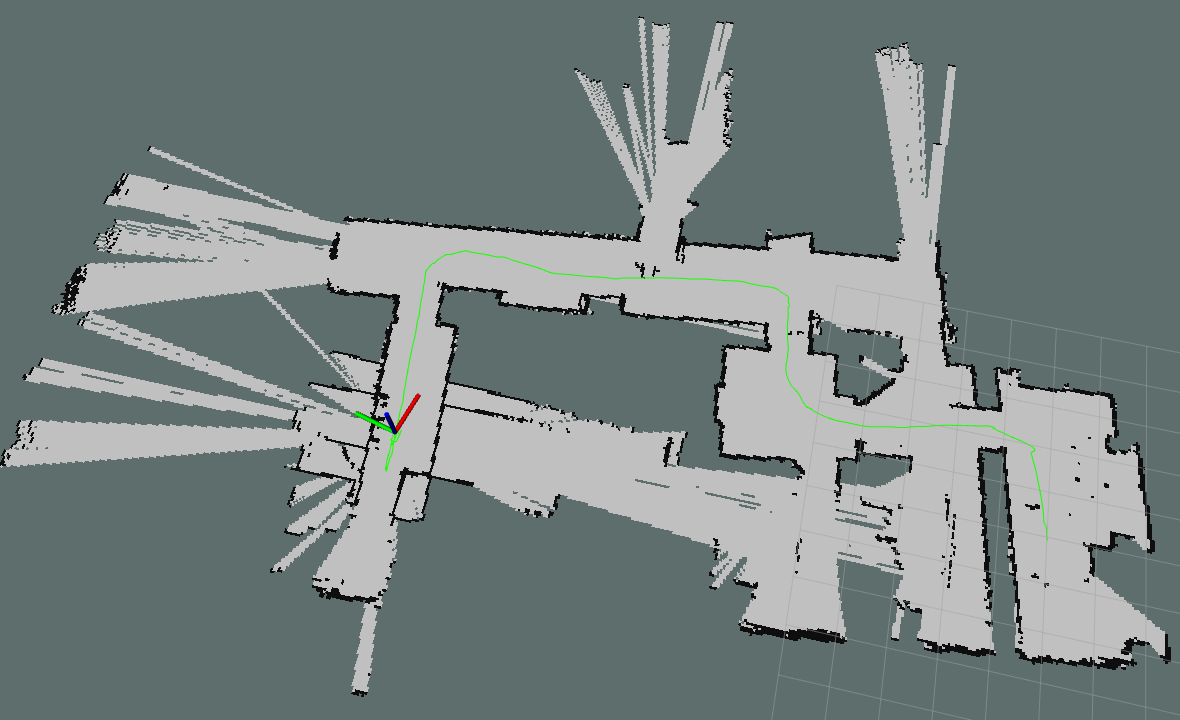}} 
    \subfigure[]{\includegraphics[width=0.51\textwidth]{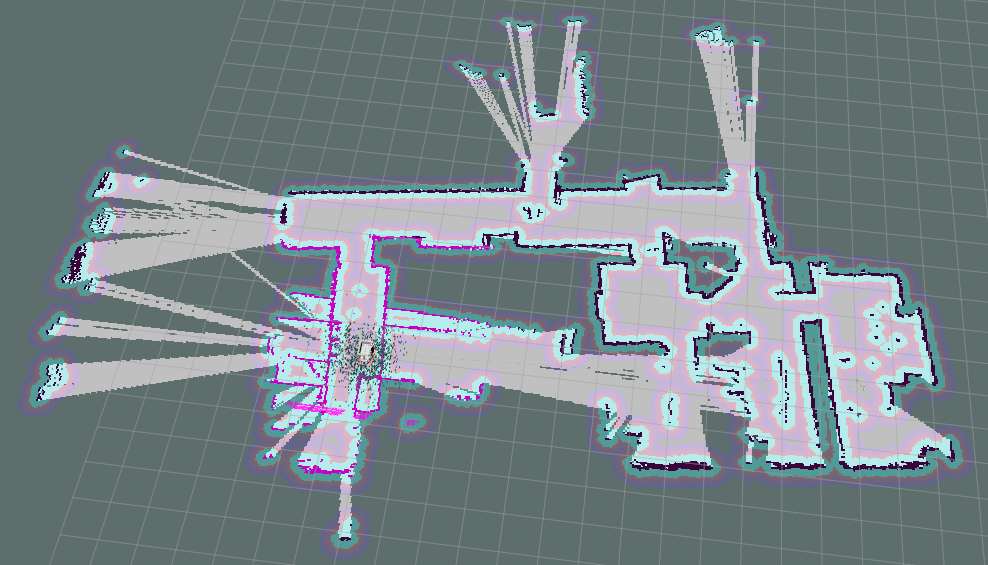}}
    \caption{Applying $ROMR$ to a real-world SLAM problem. (a) Generated the 2D occupancy grid map of the operational environment with the $P25$ lidar and the Hector-SLAM algorithm. The robot's trajectory is represented with the green line. (b) Localising the $ROMR$ within the map with the Adaptive Monte Carlo Localisation (AMCL) algorithm. The dark green clusters are the AMCL particles that represent the estimates of the location of the robot.}
    \label{fig:real_world_map}
\end{figure}

Figure \ref{fig:mapping} shows the map built in the virtual laboratory environment with the $ROMR$. Figure \ref{fig:real_world_map}a shows the map built in the real-laboratory world. For the localisation of the robot, the employed AMCL approach uses a particle filter to track the pose of the robot \cite{particlefilter}. It maintains a probability distribution over a set of all the possible robot poses \cite{amcl}, and updates this distribution using the data from the $ROMR$ odometry and rplidar scan (P25). Figure \ref{fig:real_world_map}b shows the localisation of the $ROMR$ within the 2D occupancy grid map, where the dark green clusters denote the AMCL particles representing the estimates of the location of the robot.

\subsection{Proposed maintenance for the ROMR}
Finally, to increase the life span of $ROMR$, predictive and corrective maintenance is necessary. This aims to maintain or repair the robot to ensure that the robot works at its maximum efficiency. Predictive maintenance such as listening to any abnormal noise; performing a visual inspection of all the parts of the robot; checking the energy storage level; checking for vibrations, mechanical defects, improper connections, calibration errors, etc., are proposed before each operation. This is intended to reduce the probability of failure during operation. Furthermore, curative maintenance on the other hand should be performed after detecting a failure.

\section{Conclusion}\label{sec:5}
In this paper, we presented a ROS-based open-source mobile robot $ROMR$ for research and industrial applications. We provided detailed information about the hardware design, the architecture, the operation instructions, and the advantages it offers compared to the commercial platforms. The entire design utilises off-the-shelf electronic components, additive manufacturing technologies and aluminium profiles that are commercially available to speed up the re-prototyping of the framework for custom or general-purpose applications. 
We implemented several control techniques that can enable a non-robotics expert to operate the robot easily and intuitively. Furthermore, we demonstrated the applicability of the $ROMR$ for logistics problems by implementing navigation, simultaneous localisation and mapping (SLAM) algorithms, which are fundamental prerequisites for autonomous robots. 
The experimental validation of the robustness and performance is illustrated in the video \url{https://osf.io/ku8ag}. Future work will focus on porting the whole platform to ROS 2.
As an open-source platform, the scientific community has been granted permission to use all the design files published at \url{https://doi.org/10.17605/OSF.IO/K83X7}. This open-source strategy supports rapid progress in the development of intelligent mobile robots and dexterous systems.

\section*{Declaration of interest}
None

\section*{CRediT Author Statement} 
\textbf{Nwankwo Linus:} Conceptualization, construction, software simulation, experimentation, and writing (original draft preparation). \\
\textbf{Fritze Clemens:} Software simulation, experimentation, and reviewing \\
\textbf{Konrad Bartsch:} Construction and CAD design\\
\textbf{Elmar Rueckert:} Supervision, validation, reviewing and editing \\

\section*{Acknowledgements}
This project has received funding from the Deutsche Forschungsgemeinschaft (DFG, German Research Foundation) No \#430054590 (TRAIN).

\bibliographystyle{bib_linus.bst}
\bibliography{references}

\end{document}